\documentclass[twoside]{article}
\usepackage{tcolorbox}
\tcbuselibrary{breakable, skins}
\usepackage[table]{xcolor}
\usepackage{makecell}  
\usepackage{PRIMEarxiv}
\usepackage{float}
\usepackage{multirow} 
\usepackage[utf8]{inputenc} 
\usepackage[T1]{fontenc}    
\usepackage{hyperref}       
\usepackage{url}            
\usepackage{booktabs}       
\usepackage{amsfonts}       
\usepackage{nicefrac}       
\usepackage{microtype}      
\usepackage{lipsum}
\usepackage{fancyhdr}       
\usepackage{graphicx}       
\graphicspath{{media/}}     
\usepackage{amsmath}
\usepackage{tabularx}
\newtcolorbox{promptbox}[1][]{
  title={\textbf{Prompt}},
  fonttitle=\bfseries\color{white},
  colbacktitle=gray!60!black,
  colback=white,
  colframe=gray!60!black,
  boxrule=0.5pt,
  arc=2pt,
  breakable,
  #1
}
\definecolor{bluerow}{RGB}{225, 240, 248}
\definecolor{yellowrow}{RGB}{253, 248, 225}
\definecolor{pinkrow}{RGB}{252, 230, 232}
\definecolor{greenrow}{RGB}{225, 245, 230}

\title{Beyond English benchmarks: clinical llm evaluation in Brazilian Portuguese
}

\author{
  Giordano de Pinho Souza  \\
  Federal University of Rio de Janeiro \\ Toronto Metropolitan University\\
  Toronto, CA\\
  \texttt{giordano@ufrj.br} \\
    \And
  Glaucia Melo \\
  Toronto Metropolitan University \\
   Toronto, CA \\
  \texttt{glaucia@torontomu.ca} \\
   \And
  Josefino Cabral Melo Lima \\
  Federal University of Rio de Janeiro \\
  Rio de Janeiro, RJ \\
  \texttt{clima@nce.ufrj} \\
     \And
  Daniel Schneider \\
  Federal University of Rio de Janeiro \\
  Rio de Janeiro, BR\\
  \texttt{schneider@ufrj.br} \\
}

\begin{document}
\maketitle

\begin{abstract}
Large Language Models are transforming the support for clinical decision and their application in real scenarios. Yet, most benchmarks are conducted in English, and cross-lingual evaluation is needed to tackle the language gaps in global access. We introduce ClinicalBr, the first bilingual benchmark for clinical decision built from real Brazilian case reports. The corpus contains 2,892 cases drawn from 28 SciELO medical journals, spanning 18 specialties, and is structured as parallel Portuguese–English pairs. Each case supports four evaluation tasks: diagnosis retrieval, differential diagnosis, exam recommendation, and treatment planning. We evaluate four models: MedGemma-27B, Sabiá-4, DeepSeek-R1, and o3-mini, across both languages. The central finding is that the Portuguese–English performance gap is task-dependent, not general. In diagnosis retrieval, English yields a consistent advantage across all models, with +7.5–12.1 accuracy points. This advantage disappears in differential diagnosis, exam recommendation, and treatment planning, where confidence intervals cross zero for most models and Portuguese completeness scores are marginally higher. Brazilian-endemic conditions proved easier than the full corpus, not harder, indicating that tropical presentations are adequately represented in current pre-training. Exam recommendation was the hardest task across all models and both languages, with F1 scores below 0.10, well below the differential diagnosis ceiling of 0.20–0.27.
\end{abstract}

\keywords{large language models \and cross-lingual evaluation \and clinical Decision Support \and medical benchmarks}

\section{Introduction}
\label{sec:introduction}
The advancement of LLMs has shown increasing capabilities in healthcare, improving natural language understanding to aid clinical decision-making \cite{zhao_survey_2023, qiu_towards_2024, singhal_large_2023}. Language models can decompose problems step by step and, on specific benchmark tasks, match or surpass clinicians. Because these models encode vast clinical knowledge, they perform well on structured clinical decision tasks such as diagnosis and question answering. However, this performance has been established almost entirely in English. The benchmarks that define the field were draw from English-language licensing exams \cite {pal_medmcqa_2022}, while most clinical care worldwide occurs in other languages, on records written in those languages.

Most existing medical benchmarks are English-language and built on licensing-exam multiple-choice questions \cite{qiu_quantifying_2025, zhu_diagnosisarena_2025, arora_healthbench_2025, zuo_medxpertqa_2025, wang_mmlu-pro_2024}. Both choices limit what such benchmarks can measure. Licensing exams test textbook knowledge against a fixed answer key, but clinical practice is local: disease prevalence, available diagnostic tests, and treatment guidelines differ by country. A model that scores well on US-derived exams has not thereby been shown to reason over the epidemiological profile and clinical constraints a physician faces elsewhere. Outside English, the benchmarks needed to test this are scarce. In the Brazilian context, a few Portuguese-language medical benchmarks have recently appeared \cite{godoy_alvorada-bench_2025, correia_class_2026, truyts_zero-shot_2026, bruneti_severino_benchmarking_2025, garcia_step_2025}, but they largely inherit the same exam-derived, multiple-choice format and so leave the central question — whether models reason over local clinical cases or only retrieve familiar facts — unanswered.

Cross-lingual evaluation is  needed because language gaps are tied to fairness and global access. Models are still trained mainly on high-resource languages, which leaves them less prepared for mid- and low-resource languages that are important in real-world care. Recent work links these gaps to broader corpus fairness problems, including geographic, cultural, and socio-economic bias \cite{jin_better_2024, wu_bridge_2026}. For these reasons, multilingual healthcare benchmarks are not just a technical nice-to-have; they are a basic requirement for valid evaluation. New benchmark efforts are explicitly motivated by the need to reflect the multilingual reality of medicine, broaden coverage beyond English and beyond narrow text-only tasks, and better judge whether medical LLMs are safe and useful across populations rather than only in English-speaking contexts \cite{qiu_towards_2024, ji_large_2025}

To adress this, we proposed ClinicalBr, a benchmark designed to evaluate the clinical capabilites of LLMs within the Brazilian Context. ClinicalBr pairs 2,892 real Brazilian case reports in Portuguese and English across four open-ended clinical tasks, diagnosis retrieval, differential diagnosis, exam recommendation, and treatment planning. Among all cases, we span more than 15 types of bodies and disorders, with 219 cases focused on Tropical Diseases. ClinicalBr is primarily focused on reflecting Brazilian cases and guidelines. We make three contributions:

\begin{itemize}
    \item We show that the cross-lingual performance gap in clinical LLMs is confined to single-answer retrieval and disappears under structured reasoning, a dissociation the prevailing English multiple-choice benchmarks cannot detect.
    \item We show this pattern is consistent with knowledge overlap rather than reasoning ability: the gap tracks how retrieval-anchored a model is, and conditions sparse in English corpora do not behave as a knowledge deficit would predict.
    \item We release ClinicalBR, a bilingual PT/EN benchmark of real Brazilian case reports spanning four open-ended clinical tasks, together with its paired-difference evaluation protocol, for measuring cross-lingual clinical competence.
\end{itemize}

\section{Related Work}
\label{sec:related_work}
\subsection{Medical Benchmarks for LLMs}

The evaluation of LLM efficacy in medicine has evolved from static knowledge testing toward the analysis of complex clinical competencies. Traditionally, performance was measured through medical licensing examinations, such as the USMLE and CHMLE, which form the basis of widely cited datasets like MedQA and MedMCQA \cite{jin_pubmedqa_2019, jin_what_2020}. This approach was consolidated by benchmark families such as MultiMedQA, which combines multiple sources of professional, research, and consumer health questions to test medical competence in a manner analogous to student assessment \cite{singhal_large_2023}. Although effective for measuring the mastery of standardized facts, these multiple-choice-based benchmarks have shown limitations, especially with the emergence of advanced reasoning models that have rendered these traditional challenges less rigorous \cite{nori_sequential_2025}.

In light of this saturation, recent research has shifted toward benchmarks that prioritize clinical reasoning in real-world scenarios, moving away from an exclusive reliance on academic exams. Consequently, platforms focused on specific subdomains have emerged—such as medical calculators and X-ray visual interpretation—alongside denser datasets like Medbullets and RareArena, which utilize broader data sources to simulate daily clinical practice \cite{chen_benchmarking_2024}. In this context, it is observed that the failure of traditional benchmarks lies in the insufficient evaluation of diagnostic capability, which requires complex reasoning over multidimensional patient records. To fill this gap, DiagnosisArena has emerged as a proposal designed to comprehensively evaluate diagnostic reasoning within simulated clinical environments \cite{zhu_diagnosisarena_2025}.

Parallel to this, medical evaluation has expanded beyond the question-and-answer (QA) format, encompassing tasks such as clinical summarization, information extraction, and predictive decision support. Resources such as MIMIC-III, MIMIC-IV, and ClinicalBench allow models to be tested on practical tasks, including treatment recommendations and report generation \cite{yan_large_2025}. Finally, it is common practice to also evaluate medical models on general benchmarks, such as MMLU-Pro, to ensure that refinement in a specialized domain does not compromise the model's general logic and reasoning capabilities \cite{wang_mmlu-pro_2024}.

\subsection{LLMs applied to Healthcare}

The integration of Large Language Models (LLMs) in the healthcare sector has resulted in a shift towards a more interactive and reasoning-based clinical AI, moving beyond traditional algorithms to dynamic systems capable of answering complex medical questions and retrieving specialized knowledge. Specialized models, such as Med-PaLM 2, have demonstrated the ability to handle professional-level medical examinations and provide detailed explanations for physicians, students, and patients \cite{singhal_large_2023, dullabh_integrating_nodate}. This capability extends to clinical decision support, where LLMs are increasingly used to suggest diagnoses, optimize treatment plans, and reason about patient cases through the analysis of longitudinal data from Electronic Health Records (EHR) \cite{qu_enhancing_2025, chen_llm-assisted_2026}. By mimicking human clinical reasoning, these models assist in transforming raw data into actionable insights, although researchers argue that their current role is better suited for supervised assistance than autonomous practice, due to ongoing concerns about factual grounding and safety \cite{paul_digitization_2023, abouelmehdi_big_2017}.

Beyond direct clinical reasoning, LLMs have become important in optimizing the administrative and documentation burdens that characterize modern medicine. One of the most successful applications involves the automated summarization of physician-patient dialogs, radiology reports, and discharge records, often producing results that expert reviewers judge to be superior to human-written notes in terms of clarity and conciseness \cite{wong_high-fidelity_2025, van_veen_adapted_2024}. Furthermore, these models serve as a bridge between unstructured clinical narratives and structured data systems, performing automated ICD coding and medical named entity recognition \cite{lin_case_2025}. This structured extraction is vital for hospital management and large-scale biomedical research, enabling the synthesis of vast amounts of literature and clinical trial data into living systematic reviews \cite{johri_evaluation_2025}.
The future trend in this area points toward multimodal medical foundation models that integrate diverse data streams, including text, medical images, and genomics \cite{jin_medagent-pro_2025, saab_advancing_2025}. These generalist systems are being developed to perform sophisticated tasks, such as the interpretation of chest X-rays and cross-modal data retrieval, effectively acting as comprehensive assistants across various medical specialties \cite{xia_mmedagent-rl_2026, zhang_alpacareinstruction-tuned_2025}. Additionally, LLMs are expanding their role into patient-facing functions, providing health education, triage support, and mental health assistance, thereby increasing accessibility to medical knowledge \cite{bedi_medhelm_2025}. However, as these models move closer to routine clinical care, the literature emphasizes that addressing ethical challenges — specifically regarding algorithmic bias, data privacy, and the potential for hallucinations — remains a prerequisite for their widespread adoption \cite{restrepo_multi-ophthalingua_2025}.

\section{Methodology}
\label{sec:methodology}
We introduce \textbf{ClinicalBr}, the first benchmark designed to evaluate the capabilities of state-of-the-art large language models (LLMs) in the Brazilian medical context. ClinicalBr is made from cases with medical diagnosis, treatment planning, and exam recommendation. Our benchmark integrates real-world clinical case scenarios collected from 28 Brazilian medical journals, ensuring that all cases reflect the country's distinct epidemiological profile, local clinical guideline adherence, and regional care recommendation practices.

The whole evaluation is built and evaluated by a two-stage pipeline:
A constructed masked cases with 2 languages from cases from the scielo open acess case reports, followed by four evaluation scoring in diagnosis accuracy, differential diagnosis, exam selection and treatment planning, applied to four chosen models (Sabiá-4, o3-mini, medgemma and deepseek-r1. The whole pipeline is sumarrized in Figure~\ref{pipeline}.

\(\Delta(\text{EN}-\text{PT})\) rather than two independent means.

\begin{table}[H]
\centering
\caption{Pipeline Attrition and Corpus Classification}
\label{dataset}
\begin{tabular}{p{0.2cm}lp{2.5cm}}
\toprule
\multicolumn{3}{l}{\textbf{A. Pipeline Attrition}} \\
& Discovered cases (unfiltered) & 7,177 \\
& Stage 1 — Year, publication type \& duplicate filter & 3,452 \\
& Stage 2 — Case type classification & 2,983 \\
& Stage 3 — Single-patient filter & 2,892 \\
& \textbf{Stage 4 — Final corpus} & \textbf{2,892} \\
\midrule
\multicolumn{3}{l}{\textbf{B. Case Type (Clinical Focus)}} \\
& Diagnosis & 853  \\
& Therapeutic / Treatment & 744 \\
& Both (Diagnosis and Treatment) & 1,304 \\
\midrule
\multicolumn{3}{l}{\textbf{C. Top DeCS Anatomy [A] (Body Systems)}} \\
& Cardiovascular System [A07] & 597 \\
& Musculoskeletal System [A02] & 525 \\
& Integumentary System (Skin, Hair, Nails) [A16] & 475 \\
& Nervous System [A08] & 294 \\
& Sense Organs [A09] & 257 \\
& Respiratory System [A04] & 230 \\
& Urogenital System [A05] & 177 \\
& Digestive System [A03] & 169 \\
\midrule
\multicolumn{3}{l}{\textbf{D. Top DeCS Disease [C] (Pathology)}} \\
& Infections [C01] & 710 \\
& Neoplasms [C04] & 559 \\
& Cardiovascular Diseases [C14] & 391 \\
& Skin and Connective Tissue Diseases [C17] & 192 \\
& Musculoskeletal Diseases [C05] & 168 \\
& Eye Diseases [C11] & 151 \\
& Nervous System Diseases [C10] & 143 \\
& Congenital, Hereditary, and Neonatal Abnormalities [C16] & 118 \\
\midrule
\multicolumn{3}{l}{\textbf{E. Tropical / Endemic}} \\
& Total tropical-tagged cases & 219 \\
& Endemic to Brazil & 124 \\
& Neglected Tropical Diseases (NTD) & 84 \\
& Arboviruses & 25 \\
\midrule
\multicolumn{3}{l}{\textbf{F. Top Tropical Diseases}} \\
& Tuberculosis & 71 \\
& Leprosy (Hansen's Disease) & 25 \\
& Toxoplasmosis & 24 \\
& Chagas Disease & 21 \\
& Dengue & 16 \\
& Paracoccidioidomycosis & 16 \\
& Cutaneous Leishmaniasis & 12 \\
& Visceral Leishmaniasis & 11 \\
& Chikungunya & 7 \\
& Schistosomiasis & 7 \\
\midrule
\multicolumn{3}{l}{\textbf{G. Masked Evaluation Inputs}} \\
& Diagnosis-only cases & 851 \\
& Treatment-only cases & 2,041 \\
& Exam-selection cases & 851 \\
& Cases with both (Ambos) & 1,303 \\
\bottomrule
\label{pipeline} 
\end{tabular}
\end{table}

\subsection{Corpus construction}

The first phase for discovering all the documents to parse was to find the ID of all the 28 medical related journals. After finding the IDs, we used the SciELO ArticleMeta API to fetch all the metadata from the papers. Article metadata was filtered to the case-report document type and year, and the full text was retrieved through Firecrawl \cite{firecrawl_firecrawl_nodate}. A 2 stage filtered then happeneded to improve the quality of the corpus. Stage 1 filtered each article as a clinical case report related to the diagnosis or therapeutic process with the Prompt \ref{prompt1:relevance_filter}. As we screened all the cases, multi-patient cases were identified. To prevent the inclusion of cases that could mislead the results of the LLMs, we did the Stage 2 of filtering using an LLM with the Prompt \ref{prompt:case_classifier}.  Afterall, all cases were classified in thier predoment type between diagnosis, treatment and diagnosis-treatment (both) with the Prompt \ref{prompt2:case_classifier} in the end we had 853 diagnosis cases, 744 therapeutic cases and 1,304.  Per-stage counts and the type distribution of the resulting corpus are
in Table~\ref{pipeline}.

Every filtered case was parsed into a structured JSON schema patient case that follows 3 main structures. Diagnosis related were made with Prompt  \ref{prompt6:diagnosis_masking}
, treatment related were made with Prompt \ref{prompt7:treatment_masking} and for Diagnosis and Treatment cases, the structure were made with Prompt \ref{prompt8:joint_extraction}

The model-facing version of each case masks the diagnostic answer by
stripping the entire diagnosis section and, where applicable, the
final-diagnosis field within the "presentation" field. Because the gold label
or its acronyms occasionally surfaces inside "presentation" and
"exams" (e.g.~titles, framing sentences, or verdict clauses around
investigations), we conducted a leakege verification that matches of the gold label and its acronym
variants with \texttt{{[}REDIGIDO{]}}, and additionally drops verdict clauses
gated on a long gold-token co-occurrence.

\subsection{Model panel}\label{model-panel-and-inference-protocol}

\begin{itemize}
    \item \textbf{MedGemma-27B} \cite{sellergren_medgemma_2026}, is the text-only, 27-billion-parameter variant of Google's open MedGemma collection, a dense decoder-only transformer built on the Gemma-3 backbone whose language component is further trained on a broad corpus of medical text and question–answer data. MedGemma is a collection of Gemma 3 variants trained for performance on medical text and image comprehension, making it our domain-specialised, English-centric reference point. 
    \item  \textbf{Sabiá-4}\cite{laitz_sabia-4_2026}, is the latest closed-source model in Maritaca AI's Sabiá family of Brazilian-Portuguese LLMs. It is a Transformer-based system whose exact parameter count is not publicly disclosed, developed through a multi-stage pipeline of continued pre-training on Portuguese and Brazilian legal corpora, long-context extension, instruction tuning, and preference alignment. 
    \item \textbf{DeepSeek-R1} \cite{deepseek-ai_deepseek-r1_2025}, is a 671-billion-parameter mixture-of-experts transformer post-trained with large-scale rule-based reinforcement learning to elicit explicit reasoning.
    \item \textbf{o3-mini} \cite{noauthor_introducing_nodate}, is a closed-source reasoning model from OpenAI's o-series, included as a frontier closed-weight reasoning baseline. I
\end{itemize}

\subsection{Diagnosis Accuracy}

For the diagnosis task, each benchmark model receives a masked case
consisting of only with the "presentation" and "exam" sections and is
asked to return the most likely diagnosis considering the full case. Each prediction is then paired with the reference diagnsosis from the original case.  The evaluation is done by a LLM-as-judge by \texttt{gpt-oss-120b} with the Prompt \ref{prompt15:diagnosis_grader} that graded binary if it's correct or incorrect with a short justification. The reported metric is the mean accuracy and the headline quantity is
\(\Delta(\text{EN}-\text{PT})\) with a paired per-case bootstrap \(95\%\). Diagnosis was scored on \(n{=}2{,}135\) cases paired across
both languages.

\subsection{Differential diagnosis and Exam Recommendation}

For the differential diagnosis task, each benchmark model receives a masked case "diagnosis" and is asked to return The differential-diagnosis with the prompt \ref{prompt:differential_diagnosis}. This prompt elicits a differential diagnoses --- initial candidate list, systematic
evaluation against the case, step-by-step exclusion, and a justified
final diagnosis --- and returns the candidate list as a structured array.

 The exam-recommendation task asks each model to recommend the diagnostic work-up for a masked case using the prompt \ref{prompt12:exam_selection}. Acting as the ordering physician, the model lists the complementary tests it would request — constrained to be specific (e.g., "abdominal CT with contrast" rather than "imaging") and justified by the clinical picture, and returns them as a structured object grouped into four categories: laboratory, imaging, pathology, and other.

The scoring proceeds by entity-linking both the model and the gold lists
to the UMLS Metathesaurus \cite{noauthor_unified_nodate}. Each diagnosis string is
embedded with SapBERT-XLMR \cite{liu_learning_2021}, a multilingual
SAP-tuned encoder that places PT and EN surface forms in a common
biomedical space, so EN predictions can be linked and matched by inner product against a FAISS-CPU \texttt{IndexFlatIP} index restricted to disease semantic types. Precision, Recall and F1 scores are computed over the resulting CUI sets,
with depth-\(2\) hierarchical credit: a predicted CUI counts as a hit if
it is in the gold set, an ancestor of a gold CUI, or a descendant of a
gold CUI under the UMLS CHD/RN relations.

\subsection{Treatment}\label{task-4-treatment-llm-as-judge-composite}

For the treatment task, each model receives a masked output and is
asked to write a treatment plan in free-form.  We selected 3 different judges
 ( Qwen-2.5-72B, LLaMA 3.3 70B and Gpt-oss-120b) was chosen to represent
3 types of model architechture and training, in order to decrease the bias for any given judgement. Every judge is shown the case, the proposed treatment, and the reference treatment, and returns three \(1\)--\(5\) scores: 

\begin{itemize}
    \item \textbf{Accuracy}: whether the proposed interventions are clinically correct given the diagnosis and case context;
    \item \textbf{Completeness}: whether the response covers the components present in the gold treatment
    \item \textbf{Clarity}: whether the response is written in a clinically usable, standard format. 
\end{itemize}

The composite score reported is the mean of the three axes, divided by \(5\) so that it sits on a \([0, 1]\) scale comparable to the other tasks.

\subsection{Data Leakage}

The pre-training of all the models is extremelly broad, spanning across all the years that the related cases were published, what can cause the model to memorize the answers and achieve higher scores in the benchmark. To tackle this limitation, we conducted a standard year stratification test. For each model and language, we computed the main diagnosis accuracy for each year, between 2015 and 2024. Every accuracy measure is uniform across the four models.

\begin{figure}[H]
    \centering
    \includegraphics[width=1\linewidth]{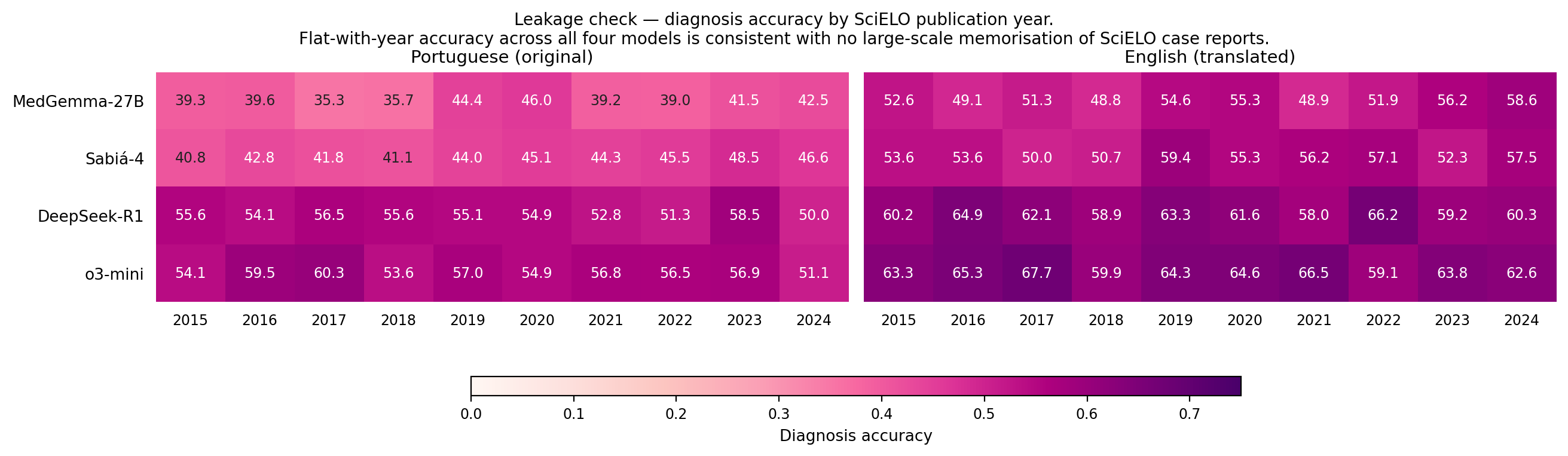}
    \caption{Accuracy in All cases}
    \label{data_leakege}
\end{figure}

\section{Results}
\label{sec:results}


\subsection{The Benchmark}


ClinicalBr is a Brazilian Portuguese clinical decision benchmark built from the SciELO Online Library. This library covers open-access academic literature from countries across South America. In this work, we use the SciELO Brazil collection of Medicine Journals, which consists of several case-report-type papers containing real information about how real clinical cases were developed. The primary language of the papers is Portuguese, however, some journals are English-based, even when reporting Brazilian cases.

The benchmark covers four stages of the clinical reasoning process: differential diagnosis, diagnosis retrieval, exam recommendation, and treatment planning. Each case has a complete version and a masked version according to the evaluation settings. For example, when evaluating diagnosis, the "diagnosis" section is removed from the structured case. Additionally, all cases are structured as parallel pairs in Portuguese and English to enable a clear cross-lingual comparison, the translation process is detailed in the methodology section. To construct the benchmark, we employ a six-step pipeline that creates a final corpus of 2,892 cases from the SciELO medical journals, spanning more than 18 medical specialties. The corpus undergoes five filtering stages to ensure its coherence and reliability within the medical domain.


\subsubsection{Data Composition}

For each case in our benchmark, there is a structured file with four fields extracted from the original case-report-type paper. The body of the file is represented by: "presentation" (e.g, clinical data, patient data, family history, past medication), "exams" (e.g, laboratory, imaging), "diagnosis" (the accurate final diagnosis of the case), "treatment" (The treatment plan with their outcomes and intervention). Considering all of the cases, not all of them are focused on diagnosis or treatment, some have only one focused area. For the final evaluation setting the benchmark had 853 diagnosis-only cases, 744 treatment-only cases, and 1,303 cases tagged "both" (both diagnostic and therapeutic focused). Portuguese and English cases are equal by construction, the pipeline was the same, preserving all the information in the four-field structure (Table \ref{dataset})

\subsubsection{Evaluation settings}

The benchmark is designed to be a open-ended questioning reasoning for the models. There are no closed questions. Each case is accompanied by the answer extracted in the original paper, which is labeled as the reference answer. In some cases, the reference answer is clear and direct, such as in the diagnosis retrieval, in other cases, the reference answer is used as a support to compare with the model predicted answer. The rationale is that in healthcare, treatment and exam recommendation rarely has a single correct answer. All evaluation is conducted in two separate language settings, Portuguese and english. A second evaluation is done specific to tropical countries such as Brazil. Table~\ref{main-results}. The following are the specifications for each type of evaluation. 

\begin{itemize}
\item \textbf{Diagnosis retrieval:} The model receives a masked case, only the clinical presentation and exam results, with diagnosis and treatment removed. The reference answer is the diagnosis stated in the original paper. Performance is measured by accuracy against that reference.
\item \textbf{Differential diagnosis:} The model receives the same masked case. The reference answer is a differential diagnosis list generated by a model that had access to the full case. Performance is measured by comparison to that list.
\item \textbf{Exam recommendation:} The model receives the same masked case. The reference answer is the set of exams documented in the full case. Each predicted exam is mapped to a UMLS CUI using SapBERT-XLMR, and performance is measured by F1 against the reference set.
\item \textbf{Treatment planning:} The model receives the masked case with the diagnosis revealed and treatment removed. The reference answer is the treatment described in the original paper. Performance is scored by an LLM-as-judge on three rubric metrics: completeness, accuracy, and clarity.
\end{itemize}

\subsubsection{Evaluation metrics}

The model's performance is evaluated through three domain-specific tasks. First, \textbf{Diagnosis Accuracy} is assessed as the accuracy of the predicted answer against the reference answer in the extracted diagnostic field, with a LLM-as-judge. The judge verifies for semantically similar diagnoses and scores with 1 or 0. Second, for Differential Diagnosis and Exam recommendation, we used feature extraction on the generated answer and match them with their UMLS CUI, then we matched the reference answer with the UMLS CUI. We then compare with 3 metrics: precision, recall, and F1. 

Finally, for Treatment Planning, a mean of three LLM-as-judge scores spanning 1–5 on a Likert scale (accuracy, completeness, and clarity) is normalized to a [0,1] scale. Treatment Accuracy measures the factual correctness of recommended elements, acting as a precision metric when comparing to the reference answer . Treatment Completeness functions as a recall metric for the treatment space, comparing to the ground truth response. Treatment Clarity evaluates the clarity and organization of the output for usage in clinical contexts. For this rubric we used the prompt \ref{prompt16:treatment_grader}

\subsection{Diagnosis: english language improve accuracy}

Table~\ref{main-results} quantify
the diagnostic advantage of English over Portuguese. Across $n{=}2{,}135$
paired masked cases, all four models score higher in English: MedGemma-27B
$+0.121$ accuracy $[+0.099, +0.143]$, Sabiá-4 $+0.107$ $[+0.086, +0.128]$,
DeepSeek-R1 $+0.075$ $[+0.053, +0.096]$, and o3-mini $+0.081$
$[+0.061, +0.104]$. Every confidence interval excludes zero. The advantage
is largest for the most language-anchored models (MedGemma, Sabiá-4) and
smallest for the strongest reasoners (DeepSeek-R1, o3-mini), consistent
with the hypothesis that the gap reflects lexical/knowledge overlap with
predominantly English biomedical pre-training rather than reasoning ability.

\begin{tabular}{llrlll}
\toprule
Task & Model & $n$ & PT (95\% CI) & EN (95\% CI) & $\Delta$ EN$-$PT (95\% CI) \\
\midrule
Diagnosis & MedGemma-27B & 2135 & 0.401 [0.381, 0.422] & 0.523 [0.501, 0.543] & 0.121 [0.099, 0.143] \\
 & Sabi\'a-4 & 2135 & 0.434 [0.412, 0.454] & 0.541 [0.518, 0.562] & 0.107 [0.086, 0.128] \\
 & DeepSeek-R1 & 2135 & 0.541 [0.521, 0.563] & 0.615 [0.594, 0.637] & 0.075 [0.053, 0.096] \\
 & o3-mini & 2135 & 0.558 [0.537, 0.579] & 0.639 [0.618, 0.660] & 0.081 [0.061, 0.104] \\
\makecell[l]{Diff diagnosis} & MedGemma-27B & 2117 & 0.205 [0.198, 0.212] & 0.229 [0.221, 0.237] & 0.024 [0.016, 0.032] \\
 & Sabi\'a-4 & 2124 & 0.218 [0.210, 0.225] & 0.214 [0.206, 0.222] & -0.003 [-0.011, 0.005] \\
 & DeepSeek-R1 & 2128 & 0.267 [0.258, 0.275] & 0.250 [0.241, 0.258] & -0.017 [-0.025, -0.009] \\
 & o3-mini & 2038 & 0.243 [0.234, 0.252] & 0.238 [0.230, 0.246] & -0.005 [-0.013, 0.003] \\
Exam selection & MedGemma-27B & 842 & 0.039 [0.033, 0.046] & 0.027 [0.022, 0.033] & -0.012 [-0.019, -0.005] \\
 & Sabi\'a-4 & 842 & 0.044 [0.038, 0.049] & 0.032 [0.027, 0.037] & -0.012 [-0.019, -0.005] \\
 & DeepSeek-R1 & 842 & 0.053 [0.047, 0.060] & 0.040 [0.035, 0.045] & -0.013 [-0.020, -0.007] \\
 & o3-mini & 842 & 0.019 [0.014, 0.025] & 0.015 [0.011, 0.019] & -0.004 [-0.010, 0.001] \\
Treatment & MedGemma-27B & 2024 & 0.807 [0.802, 0.811] & 0.780 [0.775, 0.784] & -0.027 [-0.032, -0.023] \\
 & Sabi\'a-4 & 2024 & 0.822 [0.818, 0.825] & 0.808 [0.805, 0.812] & -0.013 [-0.016, -0.010] \\
 & DeepSeek-R1 & 2024 & 0.836 [0.833, 0.839] & 0.814 [0.811, 0.817] & -0.022 [-0.025, -0.019] \\
 & o3-mini & 2024 & 0.837 [0.834, 0.840] & 0.821 [0.818, 0.824] & -0.016 [-0.019, -0.013] \\
\bottomrule
\label{main-results}
\end{tabular}

\subsection{Differential diagnosis}

\begin{figure}[H]
    \centering
    \includegraphics[width=0.75\linewidth]{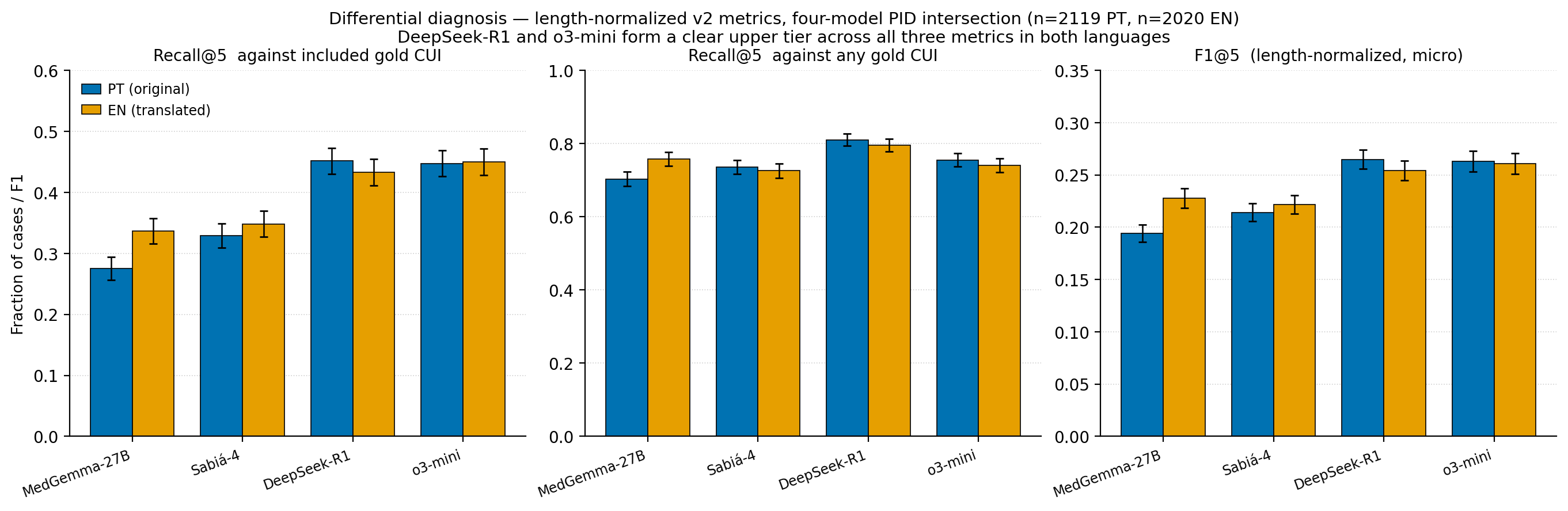}
    \caption{Comparison}
    \label{dxx}
\end{figure}

Figure \ref{dxx} shows the evaluation of the four models in the differential diagnosis evaluation of all cases. The differential diagnosis here aims to determine whether the predicted diagnoses fall within the correct space when compared to a differential diagnosis extracted from the text. We did 2 distint recall evaluation, one with the inclusion of the correct reference diagnosis and the other against any Differential Diagnosis. 

DeepSeek-R1 and o3-mini show a clear advantage. Both are the most accurate models for finding the best diagnosis, reaching $\text{Recall}@5_{\text{inc}} \approx 0.45$ in Portuguese and English, whereas Sabiá-4 and MedGemma-27B scored in the [0.28--0.34] range in the first Figure. 

The middle figure (Recall@5 against any differential Diagnosis) is much higher than the left panel (against the included gold) across all models. This indicates that the models possess a general capability to recall probable differential diagnoses in both languages; they are highly recall-driven but not necessarily precise, which is exactly why the F1@5 scores on the right collapse to the 0.2--0.27 range.

When it comes to the comparison between languages, the top-tier models perform essentially identically; DeepSeek-R1 and o3-mini are language-invariant on this task (Figure \ref{ddx_forest}). In contrast, MedGemma-27B is the only model with a visible performance gap from PT to EN, which highlights the benefits of language alignment consistent with it being an English-centric medical model.

\begin{figure}[H]
    \centering
    \includegraphics[width=0.5\linewidth]{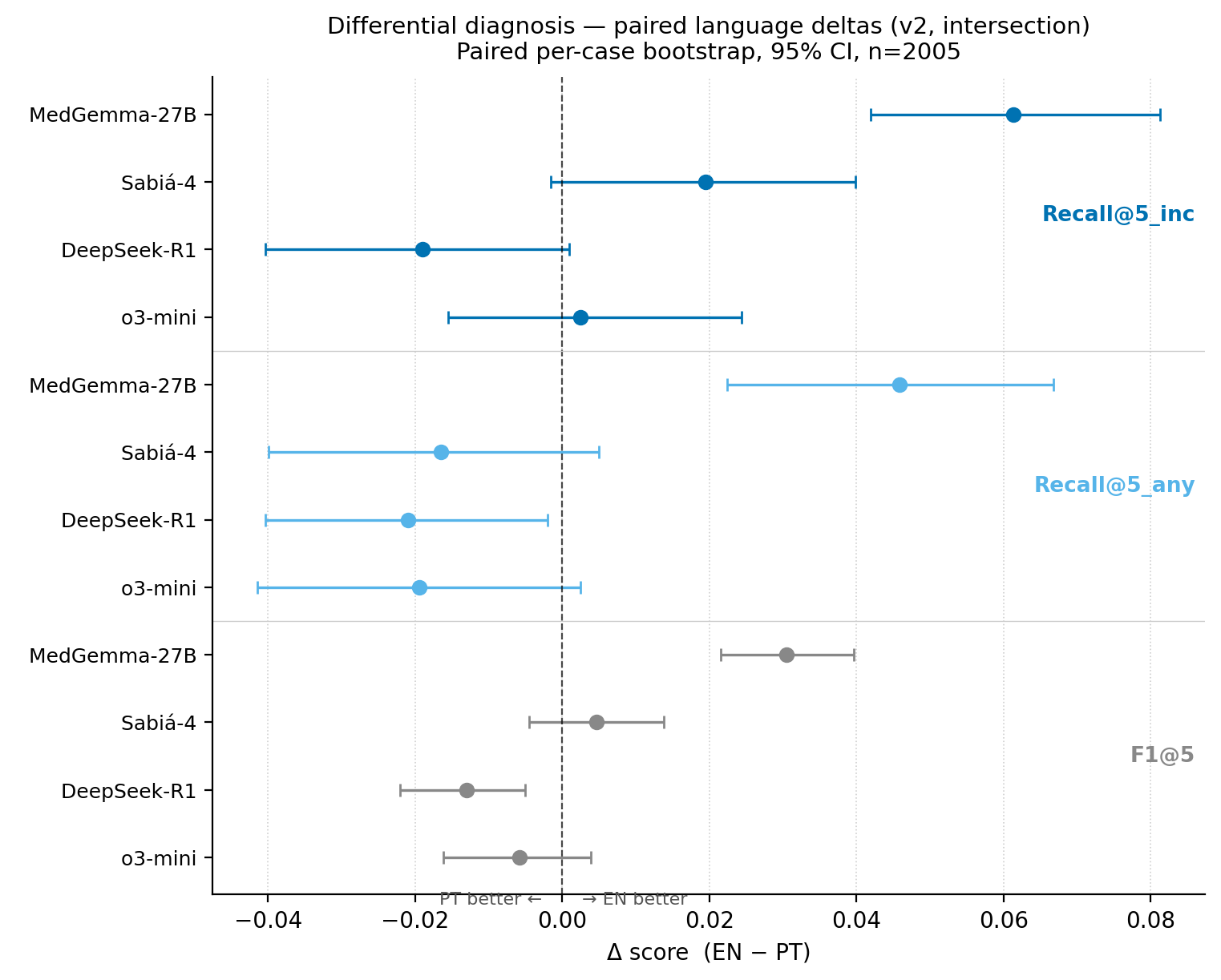}
    \caption{Comparison}
    \label{ddx_forest}
\end{figure}

\subsection{Exam Recommendation}

\begin{figure}[H]
    \centering
    \includegraphics[width=1\linewidth]{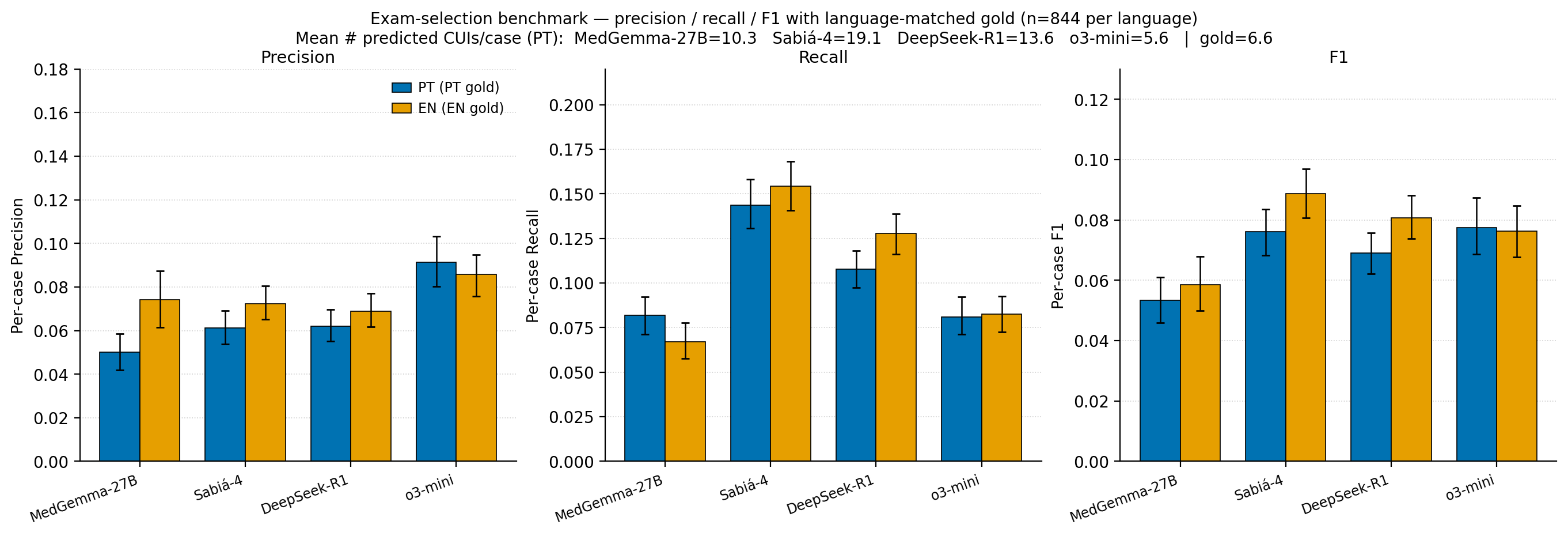}
    \caption{Precision, recall, and F1 in all exam recommendation cases}
    \label{ddx_forest}
\end{figure}

The exam recommendation evaluation is by far the worst score on all of the axis evaluated, where the maximum score is below 0.175. The read of the figure shows the precision‑recall tradeoff, with overall performance much lower than DDx.  The key differentiator between models is list length. The reference case standarts average~6.6 exams, but the models diverge widely: Sabiá-4 predicts 19.1, DeepSeek-R1 13.6, MedGemma-27B 10.3, and o3-mini 5.6, the only model within the clinically plausible range.

\begin{itemize}
    \item \textbf{Precision panel:} o3‑mini outperforms the other models (~0.09) due to its conservative approach, yielding fewer guesses and a higher hit rate. Sabiá‑4 and DeepSeek‑R1 are penalized for over‑prediction.
    \item \textbf{Recall panel:} Sabiá‑4 leads, followed closely by DeepSeek‑R1, because generating a larger list (e.g., 19 items) naturally captures more of the gold standard through higher volume. Conversely, o3‑mini's shorter list caps its recall at approximately 0.08.
    \item \textbf{F1 panel:} The two distinct strategies achieve a near-tie at ~0.08; Sabiá‑4 succeeds via output volume, while o3‑mini succeeds via selectivity, approaching the threshold from opposite directions. DeepSeek‑R1 ranks just below them, whereas MedGemma‑27B clearly performs the worst across all metrics.
\end{itemize}

F1 scores peak below 0.10. Compared to the DDx F1@5 range of 0.20–0.27, exam selection proves substantially harder than generating a diagnosis. The action space for exam recommendation is larger and comprises a wider array of options, a clinical workup involves a menu of dozens of plausible tests. Consequently, a model may select a clinically reasonable exam that simply does not match the specific choice made by the clinicians.  

Regarding language variance, English yields better results than Portuguese for three of the four models in terms of F1 score. Sabiá‑4 (0.076 $\rightarrow$ 0.088), DeepSeek‑R1 (0.070 $\rightarrow$ 0.081), and MedGemma‑27B (0.054 $\rightarrow$ 0.059) all perform better in English. Only o3‑mini demonstrates language invariance (0.077 $\approx$ 0.076). While this performance gap is modest (~1–1.5 F1 percentage points), it remains consistent across all models except o3‑mini.

\subsection{Treatment Planning}

\begin{figure}[H]
    \centering
    \includegraphics[width=0.85\linewidth]{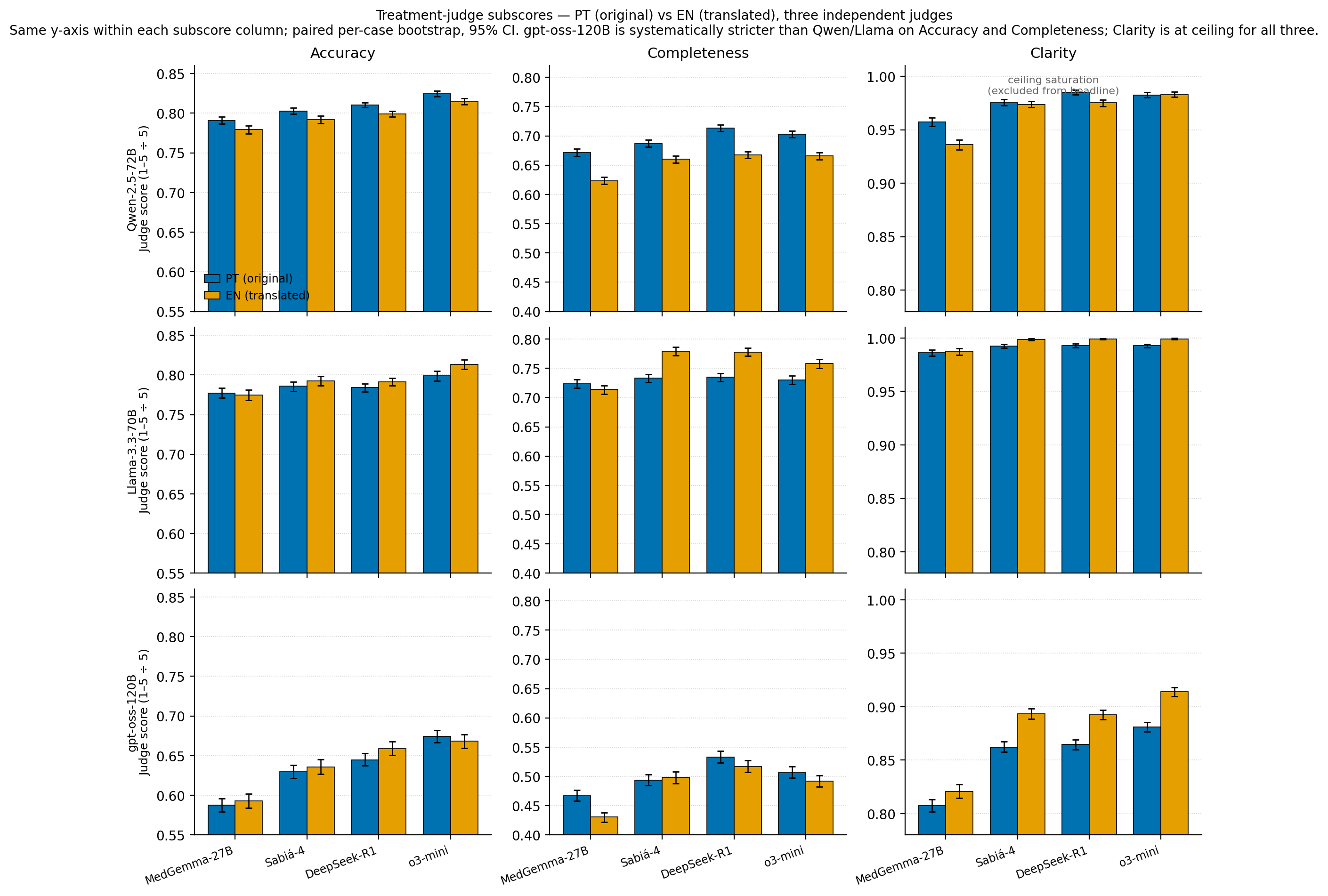}
    \caption{Treatment subscores}
    \label{treatment_subscores}
\end{figure}

The treatment evaluation constitutes the final component of our benchmark assessment. In this 
task, LLM judges evaluate open-ended responses rather than applying a binary right-or-wrong 
check. Because relying on a single judge can introduce systematic bias and variations stemming 
from model training, we employed three distinct judges: Qwen-2.5-72B, Llama-3.3-70B, and 
gpt-oss-120B. These judges scored responses across three dimensions: accuracy, completeness, 
and clarity. In total, the panel generated 16,300 scores. We present these results as a 
$3 \times 3$ grid in Figure~\ref{treatment_subscores}, from which three primary findings emerge.

First, o3-mini achieved the highest accuracy, reaching a consensus mean of 0.766. However, it fell 
short in completeness, dropping to 0.674 and ranking third behind Sabiá-4 and DeepSeek-R1. 
Consistent with our observations in the differential diagnosis task, the model that generates 
the most clinically accurate treatments is not necessarily the one that produces the most 
comprehensive plan.

Second, DeepSeek-R1 exhibited a contrasting performance profile. It scored highly in completeness, 
with a mean of 0.661, but ranked second in accuracy. It consistently generated longer treatment 
plans that incorporated diverse clinical perspectives.

Third, the clarity metric failed to differentiate the models significantly, all evaluated models 
produced clear, well-organized plans. This uniformity reflects a recurrent bias: when models 
generate longer outputs, judges tend to award higher clarity scores. Conversely, the primary 
drivers of low clarity scores were the inclusion of vague medical jargon and disorganized 
formatting. Notably, accuracy and completeness exhibited the strongest dependence on the 
reference treatment, which also corresponded to the greatest variance among judge scores. 
Disagreements between the model and the reference treatment most frequently occurred when the model 
failed to predict the identical drug or a highly relevant component of the treatment plan.

\begin{figure}[H]
    \centering
    \includegraphics[width=0.8\linewidth]{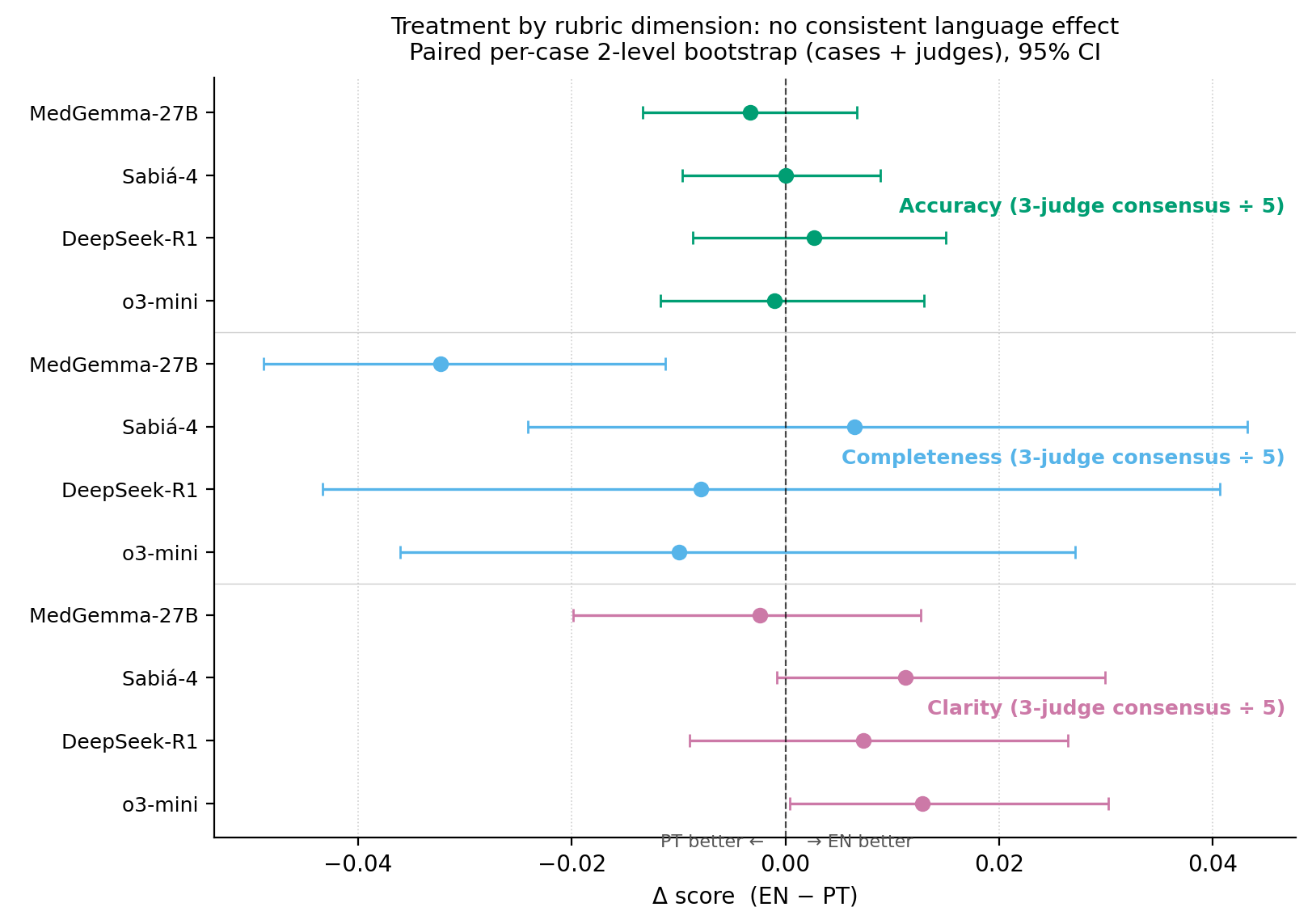}
    \caption{Treatment forest}
    \label{treatment-forest}
\end{figure}

When analyzing cross-lingual performance, accuracy scores remained closely matched, with models 
scoring between 0.72 and 0.77 in both languages. The absolute difference between English and 
Portuguese performance fell within $\pm0.005$, and every confidence interval included zero. 
For completeness, models scored between 0.62 and 0.66 in Portuguese, compared to 0.59 to 0.65 
in English. While three of the four models demonstrated a nominal advantage in Portuguese, this 
variance was not statistically significant. Ultimately, cross-checking with multiple judges 
reveals that the treatment task does not exhibit a robust Portuguese advantage. Instead, the 
language effect here is minimal, highly sensitive to the choice of LLM judge, and confined to a 
single specific criterion model combination rather than representing a systemic pattern. 
Figure~\ref{treatment-forest} illustrates this trend: eight of the treatment data points 
cluster near the zero line, with only MedGemma-27B's completeness interval leaning significantly 
toward the Portuguese side.

\subsection{Brazilian-endemic Cases}

\begin{table}[H]
  \centering
  \small
  \begin{tabular}{ll ccc ccc}
    \toprule
    & & \multicolumn{3}{c}{Diagnosis (accuracy)} & \multicolumn{3}{c}{DDx (CUI F1)} \\
    \cmidrule(lr){3-5} \cmidrule(lr){6-8}
    Model & Lang & Total & Tropical & $\Delta$ & Total & Tropical & $\Delta$ \\
    \midrule
    \multirow{2}{*}{o3-mini}
      & PT & 0.558 & 0.606 & $+0.048$ & 0.236 & 0.257 & $+0.021$ \\
      & EN & 0.639 & 0.648 & $+0.008$ & 0.231 & 0.234 & $+0.002$ \\
    \addlinespace
    \multirow{2}{*}{DeepSeek-R1}
      & PT & 0.541 & 0.601 & $+0.061$ & 0.263 & 0.275 & $+0.013$ \\
      & EN & 0.615 & 0.642 & $+0.027$ & 0.245 & 0.240 & $-0.004$ \\
    \addlinespace
    \multirow{2}{*}{Sabi\'a-4}
      & PT & 0.434 & 0.513 & $\mathbf{+0.079}$ & 0.214 & 0.250 & $\mathbf{+0.036}$ \\
      & EN & 0.541 & 0.580 & $+0.039$ & 0.212 & 0.230 & $+0.018$ \\
    \addlinespace
    \multirow{2}{*}{MedGemma-27B}
      & PT & 0.401 & 0.446 & $+0.044$ & 0.201 & 0.228 & $\mathbf{+0.027}$ \\
      & EN & 0.523 & 0.601 & $\mathbf{+0.078}$ & 0.225 & 0.255 & $\mathbf{+0.030}$ \\
    \bottomrule
  \end{tabular}
  \caption{Mean diagnosis accuracy and DDx F1 on the total corpus vs the
    219-case tropical/endemic subset.}
  \label{tab:tropical-vs-total}
\end{table}

\begin{figure}[H]
    \centering
    \includegraphics[width=1\linewidth]{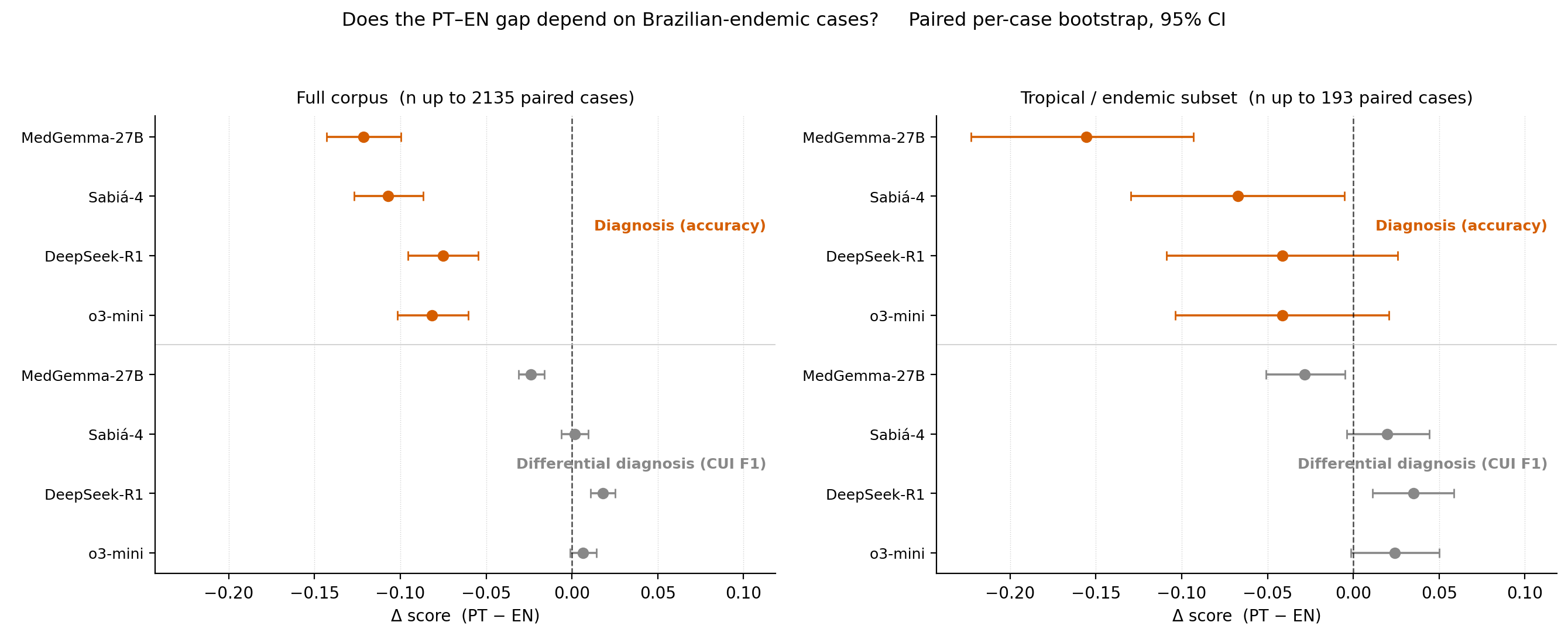}
    \caption{Tropical Forest}
    \label{tropical_forest}
\end{figure}

We next asked whether Brazilian-endemic conditions, in which English biomedical training data are sparsest, behave differently from the corpus as a whole. Restricting the evaluation to the 219 SciELO cases tagged as tropical desiese from Brazil (Table \ref{tab:tropical-vs-total}), every model–language pair scored at least as well on the tropical subset as on the full corpus, with the largest lifts in Portuguese diagnosis accuracy (+7.9+7.9 +7.9 percentage points for Sabiá-4, +6.1+6.1 +6.1 for DeepSeek-R1, and +4.8+4.8 +4.8 for GPT-o3-mini). Tropical cases were therefore systematically easier, consistent with their presentations (dengue, leishmaniasis, Chagas, schistosomiasis) being well represented in pre-training. When it comes to the PT–EN difference forest plot, Fig. \ref{tropical_forest} shows that there is no clear advantage with respect to language. Most confidence intervals overlap zero, except for MedGemma, which is mostly fine-tuned on English corpora.

\section{Discussion}
\label{sec:discussion}

\textbf{Cross-lingual gaps between language models can become a safety concern when they are utilized in a clinical decision support tool.} Language models are used daily to provide accessible care information and support for users and healthcare workers, but when it comes to their implementation in critical scenarios, language performance needs to be evaluated. Our results showed that across all four models, English language diagnosis retrieval was better than Portuguese by a margin of more than 10\% in most cases (MedGemma-27B +0.121 [+0.099, +0.143]; Sabiá-4 +0.107 [+0.086, +0.128]; DeepSeek-R1 +0.075 [+0.053, +0.096]; o3-mini +0.081 [+0.061, +0.104]), and the confidence interval excluded zero in every case. This gap was the largest for the more training language-anchored models, MedGemma-27B and Sabiá-4, and the gap was smallest for the reasoning models, DeepSeek-R1 and o3-mini. This raises a question: most hospitals and care institutions only have data in their own language about their patients, and translation can be a hard task, what does that mean for the millions of patients across the country?

\textbf{The cross-lingual difference stops when the task is not only accuracy.} When evaluation changed from simple accuracy to other metrics, such as Recall and F1, there was no longer a practical English advantage. For differential diagnosis reasoning, the difference between the languages' confidence interval crossed zero in 3 of 4 models. When it comes to exam selection, the results showed a slight advantage across some models for Portuguese. Reading again the first diagnosis claim, when it comes to a precise final answer, the English language has a clear advantage that does not generalize to tasks that require more reasoning and recalling more than one correct answer.

\textbf{Models can recall the right diagnosis without being able to prioritize them.} As stated in the last paragraph, all models perform well at gathering the most probable answers regardless of language. In the differential diagnosis evaluation, recall ranked against the suggested answer was high, in the range of 0.8, but when it comes to ranking with the correct diagnosis first, the F1 drops to 0.20–0.27. This matters for deployment in real clinical settings: the model has a great ability to recall all the possible answers, but the human-in-the-loop factor is still imperative to maintain safety for the final clinical decision. On top of that, reasoning models tend to be more precise in terms of the correct final diagnosis.

\textbf{Using AI to recommend medical tests is where current models are the furthest away from actually being useful for healthcare teams.} The F1 score never exceeded 0.1 among all of the models in any language, less than half of the differential diagnosis ceiling. Against an extracted standard from the cases, with an average of 6.6 exams, the models that proposed a higher number of exams tended to score higher. Sabiá-4 proposed 19.1, DeepSeek-R1 13.6, and MedGemma-27B 10.3, while only o3-mini stayed in the clinically plausible range at 5.6. o3-mini performed best on precision by ordering conservatively. In a real clinical scenario, particularly in underdeveloped countries, more exams means more money the patient would have to spend; in these cases, models that recommend more exams increase the cognitive load of the practitioner and can decrease their efficiency.

\textbf{Brazilian-endemic conditions were easier, not harder, for models.} We anticipated that tropical and neglected diseases that are less represented in English-centric corpora would expose more errors in the clinical decision benchmark. However, the opposite held: every model in both languages scored as well on the 219-case subset as on the full corpus. The largest gains were in Portuguese diagnosis accuracy (+7.9 points for Sabiá-4, +6.1 for DeepSeek-R1, +4.8 for o3-mini). Tropical diseases such as dengue, leishmaniasis, and Chagas are well documented in global English and biomedical text, so they are well represented in pre-training.

\textbf{The treatment-task language difference is an artifact of the judge, not a property of the models.} All treatment evaluation used three judges. Under Qwen, Portuguese plans scored higher, but this did not survive the switch to Llama-3.3-70B or the stricter gpt-oss-120B. Pooled into a consensus, the gap collapses: accuracy stays within ±0.005 between languages with every confidence interval crossing zero, and the small Portuguese edge in completeness (0.62–0.66 vs. 0.59–0.65) is not significant for any model. As Figure 7 shows, the treatment deltas cluster on the zero line, with only MedGemma-27B's completeness leaning toward Portuguese, and as the English-centric model, it is the least likely to drive a true Portuguese advantage, confirming this residual as judge variance. The implication is methodological: on open-ended tasks, a single-judge protocol would have reported a Portuguese advantage that a multi-judge protocol reveals as an artifact of one grader's scoring tendencies.

\section{Conclusion}
\label{sec:conclusion}
We introduce ClinicalBr, the first bilingual benchmark for clinical reasoning built from real Brazilian Case Reports. Spanning across more than eighteen specialties and structured as Portuguese and English pairs, it evaluated frontier models on diagnosis retrieval, differential diagnosis, exam recommendation and treatment planning under a design framework that distinguishes case difficulty from the language effect.

Our central finding is that the difference between the languages is a matter of knowledge retrieval and not reasoning of cases. English is the language that improves the best answers for single accuracy on diagnosis. Weaker models benefit the most from switching to English because they rely more on retrieval when compared to reasoning models that rely on the reasoning process before the answer. Yet, this advantage disappears the moment the task shifts to more structured reasoning for recommendations in the case. Differential diagnosis, exam selection, and treatment planning show Portuguese to be at least the equal of English, and Brazilian-endemic conditions proved easier rather than harder despite being the content least represented in English-centric corpora.

\section*{Acknowledgments}
This work was supported in part by Mitacs and the Coordenação de Aperfeiçoamento de Pessoal de Nível Superior (CAPES).

Code and data will be made publicly available upon acceptance.

\bibliographystyle{unsrt}  
\bibliography{references}  

\clearpage
\section{Suplementary}

\subsection{Qualitative Case Study}

In this section, we are gonna walk one real case through each of the four tasks,
contrasting the two strongest reasoners, DeepSeek-R1 and o3-mini, against the ground truth extracted from the source
report. Each case is drawn directly from the corpus, and every cell is reproduced verbatim from the records: the case
description is the structured case as stored in our corpus, and the model outputs are the exact predictions and judge
verdicts scored in Section 4. 

How each table is structured. Every case is laid out as one table (Tables 4–7) with a narrow row-label column on the
left. The upper block holds the shared reference answer in rows, which span the full width of the table because they are
common to both models. The lower block opens with a two-column band — DeepSeek-R1 on the left, o3-mini on the right —
under which each subsequent row places the two models' outputs side by side for the same field, so a reader can compare
them line for line. The shared reference answer in rows are:

\begin{itemize}
    \item \textbf{Case ID:} the SciELO article identifier of the case report.
    \item \textbf{Category:} the case classification — DeCS body-system [A] and disease [C] axes, plus whether the case is flagged tropical/endemic to Brazil and, if so, the matched program and disease.
    \item \textbf{Case Summary:} the full structured presentation block of the case, verbatim and field by field — patient information, chief complaint, history of present illness, past medical history, social and family history, physical examination, and hospital course (empty fields are omitted). The masked answer (the diagnosis the model must produce) does not appear here.
    \item \textbf{Final Diagnosis:} the gold diagnosis stated in the original report.
    \item \textbf{Differential Diagnosis (differential-diagnosis case only):} the gold candidate list (the Portuguese reference differential of Section 3.4); the prefix "[correct]" marks the gold-confirmed entry, and the item count is given.
    \item \textbf{Ancillary Tests Split (exam case only):} the ground-truth work-up, split into laboratory, imaging, pathology, and other, with the total exam count.
    \item \textbf{Treatment Planning Analysis (treatment case only):} the ground-truth treatment rationale, plan, and recorded outcome from the report.
\end{itemize}

\begin{table}[ht]\centering\small
\setlength{\extrarowheight}{2pt}
\begin{tabularx}{\linewidth}{|>{\raggedright\arraybackslash}p{2.7cm}|>{\raggedright\arraybackslash}X|>{\raggedright\arraybackslash}X|}
\hline
\multicolumn{3}{|l|}{\textbf{Case 1 --- Diagnosis retrieval (Neuro-Chikungunya)}} \\ \hline
\textbf{Case ID} & \multicolumn{2}{>{\hsize=\dimexpr2\hsize+2\tabcolsep+\arrayrulewidth\relax\raggedright\arraybackslash}X|}{S0004-282X2023001000922} \\ \hline
\textbf{Category} & \multicolumn{2}{>{\hsize=\dimexpr2\hsize+2\tabcolsep+\arrayrulewidth\relax\raggedright\arraybackslash}X|}{Body system: Sistema Nervoso [A08] | Disease: Infeccoes [C01]; Doencas do Sistema Nervoso [C10] | Tropical/endemic: yes (Arbovirose: Dengue, Chikungunya, Zika)} \\ \hline
\textbf{Case Summary} & \multicolumn{2}{>{\hsize=\dimexpr2\hsize+2\tabcolsep+\arrayrulewidth\relax\raggedright\arraybackslash}X|}{\textbullet{}~\emph{Patient information}: male and female sex (50\% male), mean age 59 years (range 36-75 years), occupation not reported, demographic data: hospitalized patients in a tertiary hospital during outbreaks in the state of Rio Grande do Norte, Brazil \newline \textbullet{}~\emph{Chief complaint}: neurological involvement in the context of confirmed infection \newline \textbullet{}~\emph{History of present illness}: neurological symptoms in patients with confirmed infection, median time between symptom onset and imaging 4 weeks (range 1-6 weeks) \newline \textbullet{}~\emph{Past medical history}: exclusion criterion: prior diagnosis of neurological diseases; patients with concurrent dengue or Zika virus infection were not included \newline \textbullet{}~\emph{Physical examination}: clinical examination performed by neurologist, specific details not provided \newline \textbullet{}~\emph{Hospital course}: hospitalization in a tertiary care hospital, underwent imaging studies, clinical outcome recorded at discharge} \\ \hline
\textbf{Final Diagnosis (gold)} & \multicolumn{2}{>{\hsize=\dimexpr2\hsize+2\tabcolsep+\arrayrulewidth\relax\raggedright\arraybackslash}X|}{Neuro-Chikungunya} \\ \hline
 & \multicolumn{1}{c|}{\textbf{DeepSeek-R1}} & \multicolumn{1}{c|}{\textbf{o3-mini}} \\ \hline
\textbf{Reasoning (EN)} & The 'clock face' pattern on spinal cord MRI with peripheral T2 hyperintense lesions is pathognomonic for AFM. Outbreak context and exclusion of Zika/dengue support enterovirus etiology. Anterior nerve root enhancement in one case aligns with AFM's radiculitis component. & The presence of multifocal punctate T2 hyperintense lesions in both the brain and spinal cord, occurring a few weeks after a confirmed infection, is most consistent with a post-infectious, immune-mediated demyelinating process such as ADEM. \\ \hline
\textbf{Diagnosis (EN)} & Acute Flaccid Myelitis due to enterovirus infection & Post-infectious immune-mediated demyelination (ADEM/acute transverse myelitis spectrum) \\ \hline
\textbf{Judge verdict (EN)} & incorrect (0) & incorrect (0) \\ \hline
\textbf{DeepSeek-R1 --- PT result (cross-lingual)} & \multicolumn{2}{>{\hsize=\dimexpr2\hsize+2\tabcolsep+\arrayrulewidth\relax\raggedright\arraybackslash}X|}{Complicações neurológicas por infecção por Chikungunya (mielite e encefalite) --- judged correct (1)} \\ \hline
\end{tabularx}
\caption{}\label{tab:cs-diagnosis}
\end{table}

\begin{table}[ht]\centering\small
\setlength{\extrarowheight}{2pt}
\begin{tabularx}{\linewidth}{|>{\raggedright\arraybackslash}p{2.7cm}|>{\raggedright\arraybackslash}X|>{\raggedright\arraybackslash}X|}
\hline
\multicolumn{3}{|l|}{\textbf{Case 2 --- Differential diagnosis (Sweet syndrome)}} \\ \hline
\textbf{Case ID} & \multicolumn{2}{>{\hsize=\dimexpr2\hsize+2\tabcolsep+\arrayrulewidth\relax\raggedright\arraybackslash}X|}{S0004-27492015000100014} \\ \hline
\textbf{Category} & \multicolumn{2}{>{\hsize=\dimexpr2\hsize+2\tabcolsep+\arrayrulewidth\relax\raggedright\arraybackslash}X|}{Body system: Sistema Musculoesqueletico [A02]; Sistema Digestorio [A03] | Disease: Doencas da Pele e do Tecido Conjuntivo [C17]; Oftalmopatias [C11] | Tropical/endemic: no} \\ \hline
\textbf{Case Summary} & \multicolumn{2}{>{\hsize=\dimexpr2\hsize+2\tabcolsep+\arrayrulewidth\relax\raggedright\arraybackslash}X|}{\textbullet{}~\emph{Patient information}: female, 46 years \newline \textbullet{}~\emph{Chief complaint}: painful redness on both arms and the right hand, oral mucosal lesions, arthralgia (initial presentation); red and painful left eye (late presentation) \newline \textbullet{}~\emph{History of present illness}: Initial: 7-day history of painful redness on both arms and the right hand, oral mucosal lesions that appeared and regressed during this period, and arthralgia. Late (5 months after the initial episode): complaint of a red and painful left eye lasting 10 days, without active dermatologic lesions at the time. Pain worsened with eye movements. \newline \textbullet{}~\emph{Physical examination}: Initial physical exam: red erythematous papules and vesicles on the right hand, between the fingers, and on the bilateral distal extensor areas of the upper limbs. Late ocular exam: uncorrected visual acuity of 10/10 in both eyes. Intraocular pressure of 15 mmHg in the right eye and 16 mmHg in the left eye. Elevated and hyperemic scleral area (10 × 5 mm, with surrounding episcleral injection) on the nasal side of the left eye; peripheral ulcerative keratitis adjacent to nodular scleritis. No signs of anterior chamber inflammation. Posterior segment exam without abnormalities in the left eye; nasal retina with mild pigmentary changes in the right eye. \newline \textbullet{}~\emph{Hospital course}: The patient was admitted by the rheumatology service to investigate the etiology of the arthralgia. She was referred to multiple specialties (otolaryngology, hematology, cardiology, rheumatology, neurology) for investigation of respiratory tract infection, hematologic neoplasm, cardiac involvement, arthralgia, and central nervous system involvement.} \\ \hline
\textbf{Final Diagnosis (gold)} & \multicolumn{2}{>{\hsize=\dimexpr2\hsize+2\tabcolsep+\arrayrulewidth\relax\raggedright\arraybackslash}X|}{Sweet syndrome} \\ \hline
\textbf{Differential Diagnosis (gold list)} & \multicolumn{2}{>{\hsize=\dimexpr2\hsize+2\tabcolsep+\arrayrulewidth\relax\raggedright\arraybackslash}X|}{[correct] Síndrome de Sweet (dermatose neutrofílica febril aguda); Vasculite cutânea leucocitoclástica; Pyoderma gangrenosum; Doença de Behçet; Granulomatosis com poliangiite (Wegener); Lúpus eritematoso sistêmico; Artrite reumatoide; Infecção bacteriana ou viral sistêmica; Neoplasia hematológica (ex.: leucemia mieloide aguda); Reação a fármacos; Esclerite idiopática \;(11 items)} \\ \hline
 & \multicolumn{1}{c|}{\textbf{DeepSeek-R1}} & \multicolumn{1}{c|}{\textbf{o3-mini}} \\ \hline
\textbf{Predicted DDx} & Granulomatosis with polyangiitis (GPA); Behçet's disease; Reactive arthritis; Sweet's syndrome; Sarcoidosis \;(5 items) & Behçet's disease; Sweet syndrome (acute febrile neutrophilic dermatosis); Rheumatoid arthritis \;(3 items) \\ \hline
\textbf{Recalled gold dx?} & yes & yes \\ \hline
\end{tabularx}
\caption{}\label{tab:cs-ddx}
\end{table}

\begin{table}[ht]\centering\small
\setlength{\extrarowheight}{2pt}
\begin{tabularx}{\linewidth}{|>{\raggedright\arraybackslash}p{2.7cm}|>{\raggedright\arraybackslash}X|>{\raggedright\arraybackslash}X|}
\hline
\multicolumn{3}{|l|}{\textbf{Case 3 --- Exam recommendation (Lepromatous leprosy)}} \\ \hline
\textbf{Case ID} & \multicolumn{2}{>{\hsize=\dimexpr2\hsize+2\tabcolsep+\arrayrulewidth\relax\raggedright\arraybackslash}X|}{S0365-05962016000500673} \\ \hline
\textbf{Category} & \multicolumn{2}{>{\hsize=\dimexpr2\hsize+2\tabcolsep+\arrayrulewidth\relax\raggedright\arraybackslash}X|}{Body system: Sistema Nervoso [A08]; Sistema Tegumentar (Pele, Cabelo e Unhas) [A16] | Disease: Infeccoes [C01]; Condicoes Patologicas, Sinais e Sintomas [C23] | Tropical/endemic: yes (DTN: Hanseníase)} \\ \hline
\textbf{Case Summary} & \multicolumn{2}{>{\hsize=\dimexpr2\hsize+2\tabcolsep+\arrayrulewidth\relax\raggedright\arraybackslash}X|}{\textbullet{}~\emph{Patient information}: woman, 36 years old, born and raised in Juazeiro - BA \newline \textbullet{}~\emph{Chief complaint}: ulcers on the lower limbs \newline \textbullet{}~\emph{History of present illness}: two years ago she presented edema and paresthesia in the lower limbs, accompanied by facial infiltration and hoarseness. Two months after the initial manifestations, ulcers appeared on the lower limbs. She was seen in medical consultations and referred to the vascular surgeon, with a diagnosis of stasis ulcers. Topical treatment was maintained without clinical improvement. Due to the morbidity of the ulcerative lesions, she began receiving disability benefits. \newline \textbullet{}~\emph{Past medical history}: not mentioned \newline \textbullet{}~\emph{Social history}: not mentioned \newline \textbullet{}~\emph{Family history}: not mentioned \newline \textbullet{}~\emph{Physical examination}: hoarseness, characteristic facial infiltration involving eyebrows and ears, madarosis, edema in the hands and lower limbs, large anterior and medium-sized ulcers on both legs, desquamation of the feet. Peripheral nerves not palpable. \newline \textbullet{}~\emph{Hospital course}: not mentioned} \\ \hline
\textbf{Final Diagnosis (gold)} & \multicolumn{2}{>{\hsize=\dimexpr2\hsize+2\tabcolsep+\arrayrulewidth\relax\raggedright\arraybackslash}X|}{lepromatous leprosy} \\ \hline
\textbf{Ancillary Tests Split (gold)} & \multicolumn{2}{>{\hsize=\dimexpr2\hsize+2\tabcolsep+\arrayrulewidth\relax\raggedright\arraybackslash}X|}{\emph{Laboratory}: Hemograma; Urinálise; Função hepática; Função renal; Raspado intradérmico \newline \emph{Imaging}: Tomografia computadorizada \;(6 exams)} \\ \hline
 & \multicolumn{1}{c|}{\textbf{DeepSeek-R1}} & \multicolumn{1}{c|}{\textbf{o3-mini}} \\ \hline
\textbf{Requested exams} & \emph{Laboratory}: Complete blood count (CBC); Erythrocyte sedimentation rate (ESR); C-reactive protein (CRP); Liver function tests (ALT, AST, ALP, bilirubin); Renal function tests (creatinine, BUN); Antinuclear antibody (ANA); Anti-dsDNA antibody; Antineutrophil cytoplasmic antibodies (ANCA); Anti-PGL-1 IgM serology (leprosy); VDRL; FTA-ABS; HIV antibody test \newline \emph{Imaging}: Chest X-ray; X-ray of lower limbs (to assess for osteomyelitis) \newline \emph{Pathology}: Skin biopsy from ulcer edge for histopathology; Skin biopsy for acid-fast bacilli stain (Fite-Faraco); Mycobacterial culture; Fungal culture; Bacterial culture; PCR for Mycobacterium leprae DNA \newline \emph{Other}: Slit-skin smear for acid-fast bacilli; Laryngoscopy (to evaluate hoarseness) \;(22 exams) & \emph{Laboratory}: Slit-skin smear for acid-fast bacilli (using Fite-Faraco staining); Anti-phenolic glycolipid-1 (anti-PGL-1) antibody serology \newline \emph{Pathology}: Skin biopsy from an active lesion with histopathologic examination and Fite-Faraco stain \newline \emph{Other}: PCR testing for Mycobacterium leprae on skin tissue \;(4 exams) \\ \hline
\end{tabularx}
\caption{}\label{tab:cs-exams}
\end{table}

\begin{table}[ht]\centering\small
\setlength{\extrarowheight}{2pt}
\begin{tabularx}{\linewidth}{|>{\raggedright\arraybackslash}p{2.7cm}|>{\raggedright\arraybackslash}X|>{\raggedright\arraybackslash}X|}
\hline
\multicolumn{3}{|l|}{\textbf{Case 4 --- Treatment planning (Laryngeal amyloidosis)}} \\ \hline
\textbf{Case ID} & \multicolumn{2}{>{\hsize=\dimexpr2\hsize+2\tabcolsep+\arrayrulewidth\relax\raggedright\arraybackslash}X|}{S1808-86942015000200219} \\ \hline
\textbf{Category} & \multicolumn{2}{>{\hsize=\dimexpr2\hsize+2\tabcolsep+\arrayrulewidth\relax\raggedright\arraybackslash}X|}{Body system: Sistema Respiratorio [A04]; Sistema Digestorio [A03] | Disease: Otorrinolaringopatias [C09]; Doencas do Sistema Imunitario [C20] | Tropical/endemic: no} \\ \hline
\textbf{Case Summary} & \multicolumn{2}{>{\hsize=\dimexpr2\hsize+2\tabcolsep+\arrayrulewidth\relax\raggedright\arraybackslash}X|}{\textbullet{}~\emph{Patient information}: female sex, 35 years old, black, native and resident of São Paulo/SP \newline \textbullet{}~\emph{Chief complaint}: odynophagia (worse on the right), throat clearing, pharyngeal globus and dysphonia for four years \newline \textbullet{}~\emph{History of present illness}: symptoms of odynophagia, throat clearing, pharyngeal globus and dysphonia lasting four years, without improvement with prolonged treatment with acid-reducing medication. Denied respiratory, dyspeptic or nasal symptoms \newline \textbullet{}~\emph{Past medical history}: prolonged treatment with acid-reducing medication. Denied relevant personal history \newline \textbullet{}~\emph{Social history}: denied smoking and alcohol use \newline \textbullet{}~\emph{Family history}: denied relevant family history \newline \textbullet{}~\emph{Physical examination}: hoarse voice, oroscopy and anterior rhinoscopy without abnormalities. On telelaryngoscopy: submucosal bulge in the right vestibular fold, without mobility changes, and edema/hyperemia of the interarytenoid space \newline \textbullet{}~\emph{Hospital course}: initially hypothesized another condition, lesion excision indicated with postoperative speech therapy, performed without complications} \\ \hline
\textbf{Final Diagnosis (gold)} & \multicolumn{2}{>{\hsize=\dimexpr2\hsize+2\tabcolsep+\arrayrulewidth\relax\raggedright\arraybackslash}X|}{laryngeal amyloidosis} \\ \hline
\textbf{Treatment Planning Analysis (gold)} & \multicolumn{2}{>{\hsize=\dimexpr2\hsize+2\tabcolsep+\arrayrulewidth\relax\raggedright\arraybackslash}X|}{endoscopic surgical excision of the lesion performed, followed by postoperative speech therapy. The patient was referred for follow-up in the Rheumatology department for complete evaluation. After one year, a decision was made for semi-annual outpatient follow-up and referral back to Rheumatology. Outcome: the patient attended the outpatient clinic one year after surgery asymptomatic. [...]} \\ \hline
 & \multicolumn{1}{c|}{\textbf{DeepSeek-R1}} & \multicolumn{1}{c|}{\textbf{o3-mini}} \\ \hline
\textbf{Treatment Reasoning (EN)} & Okay, let's tackle this case. The patient is a 35-year-old woman with a four-year history of odynophagia, throat clearing, globus sensation, and dysphonia. She didn't improve with acid-reducing meds, which suggests that maybe GERD isn't the main issue here. [...] & Surgical excision remains the treatment of choice for localized laryngeal amyloidosis. In this case, the recommended approach is: \textbullet{}  Surgical Procedure: Under general anesthesia, perform microlaryngoscopy with excision of the amyloid deposit from the right vestibular fold. [...] \\ \hline
\textbf{Predicted Plan} & The patient's confirmed diagnosis of laryngeal amyloidosis requires a multidisciplinary approach focusing on symptom management, prevention of recurrence, and preservation of laryngeal function. 1. [...] & Surgical excision remains the treatment of choice for localized laryngeal amyloidosis. In this case, the recommended approach is: \textbullet{}  Surgical Procedure: Under general anesthesia, perform [...] \\ \hline
\textbf{Judge scores (3-judge mean /5)} & Accuracy 3.67, Completeness 4.67, Clarity 5.00 & Accuracy 5.00, Completeness 4.00, Clarity 5.00 \\ \hline
\end{tabularx}
\caption{}\label{tab:cs-treatment}
\end{table}

\clearpage
\section{Prompts}
\section{Supplementary Prompts}

This appendix lists, verbatim, every prompt used in the construction
and evaluation of ClinicalBr. 

\subsection{Corpus-construction prompts}

\begin{promptbox}
  \label{prompt1:relevance_filter}
  \textbf{Prompt 1. Prompt for Stage-1 case-report relevance filter
  (Sim / N\~ao).}

  \medskip
  You are an experienced physician.  I will provide the full text of
  a case report.  Please help me analyze whether the case report
  focuses primarily on the diagnostic/therapeutic process.  If the
  report focuses on diagnosis/therapeutic planning, respond with
  ``Yes''.  If the report does not primarily address the task of
  diagnosis/therapeutic planning, respond with ``No''.

  \medskip\noindent\textbf{Output format:} respond ONLY in JSON with
  the following format:
\begin{verbatim}
{"verdict": "Yes" | "No",
 "justification": "<one sentence explaining your decision>"}
\end{verbatim}

  \noindent\textbf{Full text of the case report:}
  \texttt{\{full\_text\}}.
\end{promptbox}

\begin{promptbox}
  \label{prompt2:case_classifier}
  \textbf{Prompt 2. Prompt for Stage-2 case-type classifier
  (Diagnostic / Therapeutic / Both).}

  \medskip
  You are an experienced physician.  I will provide the full text of
  a case report that has already been identified as focused on
  diagnosis and/or therapeutic planning.

  \medskip\noindent\textbf{Task:} classify the main focus of the
  case report into one of the following categories.

  \begin{description}
    \item[Diagnostic:] the central challenge of the case is reaching
      the correct diagnosis (e.g., atypical presentation, complex
      differential diagnosis, decisive ancillary tests).
    \item[Therapeutic:] the central challenge is the therapeutic
      management (e.g., treatment choice, complication management,
      treatment response).
    \item[Both:] both diagnostic reasoning and therapeutic management
      are substantially developed in the text.
  \end{description}

  \medskip\noindent\textbf{Output format:} respond ONLY in JSON with
  the following format:
\begin{verbatim}
{"type": "Diagnostic" | "Therapeutic" | "Both",
 "justification": "<one sentence explaining your classification>"}
\end{verbatim}

  \noindent\textbf{Full text of the case report:}
  \texttt{\{full\_text\}}.
\end{promptbox}

\begin{promptbox}
  \label{prompt3:anatomy_categorisation}
  \textbf{Prompt 3. Prompt for DeCS anatomy categorisation
  (Hierarchy [A]).}

  \medskip
  Please categorize the body systems and anatomical structures
  involved in the health conditions and disease phenotypes discussed
  in the following case report.  Use the full text provided to
  determine the most relevant category or categories.

  \medskip\noindent\textbf{Rules:}
  \begin{itemize}
    \item If the structure is not listed, return ``Others''.
    \item Provide the category or categories directly, without
          additional words.
    \item If there are multiple categories, separate them with
          \texttt{\textbackslash n}.
  \end{itemize}

  \medskip\noindent\textbf{Categories (DeCS Hierarchy [A]):}
  \begin{itemize}
    \item Body Structures [A01]; Musculoskeletal System [A02];
          Digestive System [A03]; Respiratory System [A04]; Urogenital
          System [A05]; Endocrine System [A06]; Cardiovascular System
          [A07]; Nervous System [A08]; Sense Organs [A09]; Tissues
          [A10]; Cells [A11]; Fluids and Secretions [A12];
          Stomatognathic System (Mouth and Teeth) [A14]; Hemic and
          Lymphatic Systems (Blood and Immunological) [A15];
          Integumentary System (Skin, Hair and Nails) [A16];
          Embryonic Structures [A17].
  \end{itemize}

  \medskip\noindent\textbf{Full text of the case report:}
  \texttt{\{full\_text\}}.
\end{promptbox}

\begin{promptbox}
  \label{prompt4:disease_categorisation}
  \textbf{Prompt 4. Prompt for DeCS disease categorisation
  (Hierarchy [C]).}

  \medskip
  Please categorize the diseases and symptoms discussed in the
  following case report.  Use the full text provided to determine the
  most relevant category or categories from the list below.

  \medskip\noindent\textbf{Rules:}
  \begin{itemize}
    \item If the diseases and symptoms do not fit any of the listed
          categories, return ``Others''.
    \item Provide the category or categories directly, without
          additional words.
    \item If there are multiple categories, separate them with
          \texttt{\textbackslash n}.
  \end{itemize}

  \medskip\noindent\textbf{Categories (DeCS Hierarchy [C]):}
  \begin{itemize}
    \item Infections [C01]; Neoplasms [C04]; Musculoskeletal Diseases
          [C05]; Digestive System Diseases [C06]; Stomatognathic
          Diseases [C07]; Respiratory Tract Diseases [C08];
          Otorhinolaryngologic Diseases [C09]; Nervous System Diseases
          [C10]; Eye Diseases [C11]; Urogenital Diseases [C12];
          Cardiovascular Diseases [C14]; Hemic and Lymphatic Diseases
          [C15]; Congenital, Hereditary, and Neonatal Diseases and
          Abnormalities [C16]; Skin and Connective Tissue Diseases
          [C17]; Nutritional and Metabolic Diseases [C18]; Endocrine
          System Diseases [C19]; Immune System Diseases [C20];
          Disorders of Environmental Origin [C21]; Animal Diseases
          [C22]; Pathological Conditions, Signs and Symptoms [C23];
          Occupational Diseases [C24]; Chemically-Induced Disorders
          [C25]; Wounds and Injuries [C26].
  \end{itemize}

  \medskip\noindent\textbf{Full text of the case report:}
  \texttt{\{full\_text\}}.
\end{promptbox}

\begin{promptbox}
  \label{prompt6:diagnosis_masking}
  \textbf{Prompt 6. Prompt for generating diagnosis-task data from a
  case report (diagnosis masking).}

  \medskip
  You are an experienced physician.  Extract the clinical
  information from the case report below into structured JSON.

  \medskip\noindent\textbf{Critical rule --- presentation without
  leakage.}  In the \texttt{presentation}, NEVER mention names of
  diseases, syndromes, or diagnoses in any form:
  \begin{itemize}
    \item ``diagnosed with X'', ``confirmed X'', ``suspected X'',
          ``highly suspicious for X'';
    \item ``considering X'', ``hypothesis of X'', ``consistent with
          X'', ``suggestive of X'';
    \item ``to investigate X'', ``to rule out X'', ``carrier of X'',
          ``history of X''.
  \end{itemize}

  \medskip\noindent\textbf{Schema --- \texttt{presentation}:}
  \begin{description}
    \item[patient\_info:] sex, age, occupation, relevant
      demographics.
    \item[chief\_complaint:] main complaint that prompted the visit.
    \item[history\_present\_illness:] detailed chronological history
      of present illness, symptoms, progression.
    \item[past\_medical\_history:] past medical history,
      comorbidities, previous surgeries.
    \item[social\_history:] lifestyle habits, smoking, alcohol,
      travel, occupational exposures.
    \item[family\_history:] relevant family history.
    \item[physical\_exam:] detailed physical examination findings.
    \item[hospital\_course:] clinical course during hospitalization,
      complications, procedures performed.
  \end{description}

  \noindent\textbf{Schema --- \texttt{workup}:}
  \begin{description}
    \item[laboratory:] CBC, biochemistry, serology, cultures,
      inflammatory markers, renal/liver function, etc.
    \item[imaging:] findings from X-ray, CT, MRI, ultrasound,
      echocardiogram, PET-scan, etc.
    \item[pathology:] biopsy results, histopathology,
      immunohistochemistry, cytology.
    \item[other:] other relevant ancillary tests --- ECG, spirometry,
      endoscopy, etc.
  \end{description}

  \noindent\textbf{Schema --- \texttt{diagnosis}:}
  \begin{description}
    \item[diagnosis:] definitive diagnosis with explanation based on
      clinical findings and tests that confirm it.
    \item[final\_diagnosis\_name:] only the name of the final
      diagnosis, no conclusions.
  \end{description}

  \noindent\textbf{Schema --- \texttt{treatment}:}
  \begin{description}
    \item[intervention:] treatment performed, if mentioned in the
      text, or \texttt{null}.
    \item[outcome:] patient outcome, if mentioned, or \texttt{null}.
    \item[follow\_up:] follow-up, if mentioned, or \texttt{null}.
  \end{description}

  \medskip\noindent
  Missing fields = \texttt{null}.  Do not fabricate data.  Keep
  medical terms in English.

  \medskip\noindent\textbf{Article text:}
  \texttt{\{full\_text\}}.
\end{promptbox}

\begin{promptbox}
  \label{prompt7:treatment_masking}
  \textbf{Prompt 7. Prompt for generating treatment-task data from a
  therapeutic case report (treatment masking).}

  \medskip
  You are an experienced physician.  Extract the clinical
  information from the therapeutic case report below into structured
  JSON.

  \medskip\noindent\textbf{Critical rule --- presentation without
  treatment leakage.}  In the \texttt{presentation}, NEVER mention
  information about the treatment performed:
  \begin{itemize}
    \item names of medications, doses, dosages;
    \item surgical or therapeutic procedures performed;
    \item treatment results or outcomes;
    \item post-treatment evolution, follow-up, prognosis.
  \end{itemize}

  The diagnosis MUST be included in the presentation
  (\texttt{final\_diagnosis} field).  The goal is for the model to
  receive the complete case with diagnosis, but without knowing which
  treatment was given.

  \medskip\noindent\textbf{Schema --- \texttt{presentation}:}
  \begin{description}
    \item[patient\_info:] sex, age, occupation, relevant
      demographics.
    \item[chief\_complaint:] main complaint that prompted the visit.
    \item[history\_present\_illness:] history of present illness
      with chronology, symptoms, progression.
    \item[past\_medical\_history:] past medical history,
      comorbidities, previous surgeries.
    \item[allergies:] known allergies, or \texttt{null}.
    \item[social\_history:] lifestyle habits, smoking, alcohol,
      travel, occupational exposures.
    \item[family\_history:] relevant family history.
    \item[physical\_exam:] detailed physical examination findings.
    \item[final\_diagnosis:] definitive diagnosis of the patient.
  \end{description}

  \noindent\textbf{Schema --- \texttt{workup}:}
  \begin{description}
    \item[laboratory:] CBC, biochemistry, serology, cultures,
      inflammatory markers, renal/liver function, etc.
    \item[imaging:] findings from X-ray, CT, MRI, ultrasound,
      echocardiogram, PET-scan, etc.
    \item[pathology:] biopsy results, histopathology,
      immunohistochemistry, cytology.
    \item[other:] other relevant ancillary tests --- ECG, spirometry,
      endoscopy, etc.
  \end{description}

  \noindent\textbf{Schema --- \texttt{treatment}:}
  \begin{description}
    \item[intervention:] detailed treatment description ---
      medications with doses, dosage and duration; surgical
      procedures with technique used; complementary therapies;
      changes in management during treatment.
    \item[outcome:] detailed outcome --- treatment response, symptom
      resolution, post-treatment complications, follow-up test
      results, patient functional status.
    \item[follow\_up:] post-treatment follow-up --- return visits,
      follow-up tests, long-term evolution, recurrences, therapeutic
      adjustments.
  \end{description}

  \medskip\noindent
  Missing fields = \texttt{null}.  Do not fabricate data.  Keep
  medical terms in English.

  \medskip\noindent\textbf{Article text:}
  \texttt{\{full\_text\}}.
\end{promptbox}

\begin{promptbox}
  \label{prompt8:joint_extraction}
  \textbf{Prompt 8. Prompt for generating joint diagnosis-and-
  treatment data (\emph{Ambos}) from a case report.}

  \medskip
  You are an experienced physician.  Extract the clinical
  information from the case report below into structured JSON, with
  depth in both diagnosis and treatment.

  \medskip\noindent\textbf{Critical rule --- presentation without
  leakage.}  In the \texttt{presentation}, NEVER mention names of
  diseases / syndromes / diagnoses NOR names of treatments /
  medications / surgeries, in any form:
  \begin{itemize}
    \item ``diagnosed with X'', ``suspected X'', ``considering X'',
          ``consistent with X'', ``carrier of X'';
    \item ``treated with X'', ``on X'', ``underwent X'', ``started
          X''.
  \end{itemize}

  \medskip\noindent\textbf{Schema --- \texttt{presentation}:}
  \begin{description}
    \item[patient\_info:] sex, age, occupation, relevant
      demographics.
    \item[chief\_complaint:] main complaint that prompted the visit.
    \item[history\_present\_illness:] detailed chronological history
      of present illness, symptoms, progression.
    \item[past\_medical\_history:] past medical history,
      comorbidities, previous surgeries.
    \item[social\_history:] lifestyle habits, smoking, alcohol,
      travel, occupational exposures.
    \item[family\_history:] relevant family history.
    \item[physical\_exam:] detailed physical examination findings.
    \item[hospital\_course:] clinical course before treatment,
      initial complications.
  \end{description}

  \noindent\textbf{Schema --- \texttt{workup}:} \texttt{laboratory},
  \texttt{imaging}, \texttt{pathology}, \texttt{other}.

  \noindent\textbf{Schema --- \texttt{diagnosis}:}
  \begin{description}
    \item[diagnosis:] definitive diagnosis with explanation based on
      clinical findings and tests that confirm it.
    \item[final\_diagnosis\_name:] only the name of the final
      diagnosis, no conclusions.
  \end{description}

  \noindent\textbf{Schema --- \texttt{treatment}:}
  \begin{description}
    \item[intervention:] detailed treatment description.
    \item[outcome:] detailed outcome.
    \item[follow\_up:] post-treatment follow-up.
  \end{description}

  \medskip\noindent
  Missing fields = \texttt{null}.  Do not fabricate data.  Keep
  medical terms in English.

  \medskip\noindent\textbf{Article text:}
  \texttt{\{full\_text\}}.
\end{promptbox}

\begin{promptbox}
  \label{prompt9:validation_audit}
  \textbf{Prompt 9. Prompt for validating the fidelity of the
  structured-case extraction (hallucination and completeness audit).}

  \medskip
  You are a medical accuracy validator for a Portuguese clinical
  benchmark.  You will receive (i) the original clinical case
  report (\texttt{\{fulltext\}}) and (ii) a structured JSON
  extraction of that case (\texttt{\{structured\_json\}}).  Verify
  the extraction fidelity against the fulltext.

  \medskip\noindent\textbf{Checks:}
  \begin{description}
    \item[Accuracy (hallucinations):] every clinical claim in the
      structured JSON must be supported by the fulltext.  Flag any
      fabricated or unsupported content (dates, labs, medications,
      findings, diagnoses that do not appear in the source).
    \item[Completeness:] important clinical details present in the
      fulltext must be represented in the structured JSON.  Flag
      meaningful omissions (key exam findings, pivotal lab values,
      relevant past medical history, treatment specifics, outcome).
    \item[Diagnosis correctness:] the
      \texttt{diagnostico.diagnostico} /
      \texttt{diagnostico.resultado\_final\_diagnostico} fields must
      match the final diagnosis stated or clearly implied in the
      fulltext.
  \end{description}

  Do NOT check ICD-10 codes.  The fulltext may be partially redacted
  for benchmark purposes (tokens like \texttt{[REDIGIDO]} are
  intentional --- do not treat them as errors).  Minor rewording,
  summarization, and translation between PT / EN are acceptable.

  \medskip\noindent\textbf{Output format} (return ONLY a JSON object;
  no prose, no markdown fences):
\begin{verbatim}
{
  "is_valid": true|false,
  "accuracy_score": 0.0-1.0,
  "completeness_score": 0.0-1.0,
  "issues": [
    {
      "field": "path to problematic field",
      "issue_type": "hallucination|missing|incorrect|incomplete",
      "description": "concise description of the issue",
      "suggested_fix": "corrected value or null"
    }
  ]
}
\end{verbatim}

  A case is \texttt{is\_valid: true} only if
  \texttt{accuracy\_score $\geq$ 0.85} AND there are no
  \texttt{hallucination} issues.  Only flag genuine problems; minor
  phrasing differences are acceptable.
\end{promptbox}

\subsection{Benchmark-inference prompts}

\begin{promptbox}
  \label{prompt11:diagnosis_reasoning}
  \textbf{Prompt 11. Prompt for diagnosis reasoning (differential
  diagnosis plus final diagnosis).}

  \medskip
  You are an experienced physician who will receive a structured
  patient case in JSON format.  Your task is to analyze all patient
  data to build a differential diagnosis and a final diagnosis.

  \medskip\noindent\textbf{Differential diagnosis process.}
  First, based on the patient's chief complaint and clinical
  information, generate an initial list of possible diseases.  This
  list will then be systematically evaluated, comparing each disease
  against the patient's symptoms, signs, and test results.  Next, the
  rationale for including or excluding each diagnosis will be
  explained based on clinical evidence, highlighting the selection
  of diagnostic tests that help confirm or rule out specific
  conditions.  The process involves step-by-step elimination of less
  likely diagnoses, culminating in the most probable diagnosis,
  which will be justified by clinical reasoning and the evidence
  supporting it.

  \medskip\noindent\textbf{Final diagnosis.}  Integrate the
  patient's clinical presentation, test results, and differential
  diagnosis process to summarize the final diagnosis.  Briefly
  explain the basis for the diagnosis and highlight the key factors
  that support this conclusion.

  \medskip\noindent\textbf{Diagnosis result.}  Present only the
  diagnostic result, without any further explanation.

  \medskip\noindent\textbf{Output format.}  Finalize your response
  with a JSON block in exactly this format:
\begin{verbatim}
{
  "differential_diagnosis": [
    {"diagnosis": "...", "included": true,
     "justification": "..."},
    {"diagnosis": "...", "included": false,
     "justification": "..."}
  ],
  "explanatory_final_diagnosis": "...",
  "diagnosis_result": "..."
}
\end{verbatim}

  \noindent\textbf{Clinical case:} \texttt{\{case\_text\}}.
\end{promptbox}

\begin{promptbox}
  \label{prompt12:exam_selection}
  \textbf{Prompt 12. Prompt for exam-selection reasoning
  (complementary work-up).}

  \medskip
  You are an experienced physician who will receive a structured
  patient case in JSON format.  Your task is to recommend all
  necessary tests to pass on to another doctor who will make the
  diagnosis.  List the complementary tests you would order now.

  \medskip\noindent\textbf{Rules:}
  \begin{itemize}
    \item Be specific (e.g., ``Abdominal CT scan with contrast'',
          not ``imaging'').
    \item Include only tests justified by the clinical picture.
  \end{itemize}

  \medskip\noindent\textbf{Output format} (return only the JSON
  object below):
\begin{verbatim}
{"laboratory": [], "imaging": [],
 "pathology":  [], "other":   []}
\end{verbatim}

  \noindent\textbf{Case:} \texttt{\{case\_json\}}.
\end{promptbox}

\begin{promptbox}
  \label{prompt13:treatment_planning}
  \textbf{Prompt 13. Prompt for treatment-planning reasoning.}

  \medskip
  You are an experienced physician reasoning about a clinical case.
  You will receive an organized excerpt from a patient's case.
  Consider all the information to complete the task.  Your task is
  to determine the most likely treatment for this case.

  \medskip\noindent\textbf{Output format.}  Wrap the answer in
  \texttt{<treatment>...</treatment>} tags containing a detailed
  description of the treatment:
  \begin{itemize}
    \item medications with doses, dosage regimen, and duration;
    \item surgical procedures with technique used;
    \item complementary therapies;
    \item changes in management during treatment.
  \end{itemize}

  \medskip\noindent\textbf{Case:} \texttt{\{case\_json\}}.
\end{promptbox}

\begin{promptbox}
  \label{prompt14:exam_canonicalisation}
  \textbf{Prompt 14. Prompt for canonicalising the model's predicted
  exams prior to UMLS entity linking.}

  \medskip
  You are a clinical annotator.  You will receive a list of
  complementary exams recommended by a model, already organized into
  four categories: \texttt{laboratory}, \texttt{imaging},
  \texttt{pathology}, \texttt{other}.  Your task is to CANONICALIZE
  each item --- not to add, remove, or reinterpret --- and return
  the same structure in canonical form.

  \medskip\noindent\textbf{Canonicalization rules:}
  \begin{itemize}
    \item Preserve every exam present.  Do not invent new exams;
          do not drop exams present in the input.
    \item If a single entry contains multiple exams, split it into
          one entry per exam.  Examples:
          ``Syphilis serology (VDRL/FTA-ABS) and HIV'' $\rightarrow$
          ``VDRL serology'', ``FTA-ABS test'', ``HIV serology'';
          ``Electroretinogram (ERG) and visual evoked potential
          (VEP)'' $\rightarrow$ ``Electroretinogram (ERG)'',
          ``Visual evoked potential (VEP)'';
          ``Renal function (urea and creatinine)'' $\rightarrow$
          ``Urea'', ``Creatinine'';
          ``Lipid panel (total cholesterol, HDL, LDL,
          triglycerides)'' $\rightarrow$ ``Total cholesterol'',
          ``HDL'', ``LDL'', ``Triglycerides''.
    \item Strip justifications, clinical indications, and any
          explanatory text in parentheses.  E.g.,
          ``High-sensitivity CRP (inflammatory marker)''
          $\rightarrow$ ``High-sensitivity CRP'';
          ``Brain MRI with contrast (evaluate leukodystrophies)''
          $\rightarrow$ ``Brain MRI with contrast''.
    \item Keep parenthetical text ONLY when it is part of the exam
          name (an acronym, a sequence, or an anatomical/technical
          modifier), not a justification.  E.g.,
          ``Erythrocyte sedimentation rate (ESR)'' --- keep;
          ``Chest CT'' --- keep;
          ``Complete blood count (to investigate anemia)''
          $\rightarrow$ ``Complete blood count''.
    \item Use the standardized English exam name.  When both an
          acronym and a full name appear, prefer the full name (keep
          the acronym in parentheses if clinically useful).
    \item Do not move exams between categories.
    \item Do not duplicate the same canonical exam within the same
          category.
    \item If a category is empty in the input, return an empty list.
  \end{itemize}

  \medskip\noindent\textbf{Output format} (return only the JSON
  block below; no comments, no text before or after):
\begin{verbatim}
{"laboratory": [], "imaging": [],
 "pathology":  [], "other":   []}
\end{verbatim}

  \noindent\textbf{Model recommendations:}
  \texttt{\{case\_json\}}.
\end{promptbox}

\subsection{LLM-as-judge prompts}

\begin{promptbox}
  \label{prompt15:diagnosis_grader}
  \textbf{Prompt 15. Prompt for the LLM-as-judge diagnosis grader
  (binary correct / incorrect).}

  \medskip
  You are a medical evaluation judge.  Your task is to compare a
  student's diagnosis response with the correct diagnosis and
  evaluate the prediction.

  \medskip\noindent\textbf{Inputs:}
  \begin{description}
    \item[Correct diagnosis:] \texttt{\{correct\_diagnosis\}}.
    \item[Student's diagnosis:] \texttt{\{model\_diagnosis\}}.
  \end{description}

  \noindent\textbf{Evaluation criteria:}
  \begin{description}
    \item[Correct:] the diagnosis is identical or clinically
      equivalent.
    \item[Incorrect:] the diagnosis is wrong or unrelated to the
      correct one.
  \end{description}

  \medskip\noindent\textbf{Output format:}
\begin{verbatim}
{
  "verdict": "correct" | "incorrect",
  "justification": "Brief explanation in 1-2 sentences."
}
\end{verbatim}
\end{promptbox}

\begin{promptbox}
  \label{prompt16:treatment_grader}
  \textbf{Prompt 16. Prompt for the LLM-as-judge treatment grader
  (accuracy / completeness / clarity, 1--5 Likert).}

  \medskip
  You are a medical expert evaluating a treatment plan proposed for
  a clinical case.  Your goal is to assess how well the proposed
  treatment addresses the patient's condition and compare it to the
  reference treatment (gold standard) in terms of accuracy,
  completeness, and clarity.

  \medskip\noindent\textbf{Inputs:}
  \begin{description}
    \item[Clinical case:] presentation, exams, and confirmed
      diagnosis, provided inside
      \texttt{<case>\{\{CASE\}\}</case>}.
    \item[Generated treatment:] provided inside
      \texttt{<treatment>\{\{TREATMENT\}\}</treatment>}.
    \item[Reference treatment:] gold standard reflecting what was
      actually done for the patient, provided inside
      \texttt{<gold\_treatment>\{\{GOLD\_TREATMENT\}\}</gold\_treatment>}.
  \end{description}

  Carefully review the \texttt{<treatment>} in light of the
  \texttt{<case>} and compare it to the
  \texttt{<gold\_treatment>}.  For each of the following criteria,
  rate the response on a scale of 1 to 5 (1 = very poor, 5 =
  excellent) and provide a short justification.

  \medskip\noindent\textbf{Evaluation criteria:}
  \begin{description}
    \item[Accuracy (1--5):] are the proposed interventions
      clinically correct for the diagnosis, case, and gold-standard
      context?
    \item[Completeness (1--5):] does the treatment cover the
      components present in the gold standard?
    \item[Clarity (1--5):] is the treatment written clearly and
      organized in a standard clinical format usable by clinicians?
  \end{description}

  \medskip\noindent\textbf{Output format.}  Output the evaluation as
  a single valid JSON object matching the following structure:
\begin{verbatim}
{
 "accuracy":    {"score":0,
                 "explanation":"Explain why this score was given."},
 "completeness":{"score":0,
                 "explanation":"Explain why this score was given."},
 "clarity":     {"score":0,
                 "explanation":"Explain why this score was given."}
}
\end{verbatim}

  \noindent\textbf{Ensure the output is valid JSON:}
  \begin{itemize}
    \item use double quotes (\texttt{"}) for all keys and string
          values;
    \item when quoting text or sections inside the explanations, use
          escaped double quotes (\texttt{\textbackslash"}) to
          maintain valid JSON formatting;
    \item do not include any additional information in the output.
  \end{itemize}
\end{promptbox}

\subsection{Translation prompts}

\begin{promptbox}
  \label{prompt17:pt_en_translation}
  \textbf{Prompt 17. Prompt for Portuguese-to-English translation of
  full-text case reports.}

  \medskip
  You are a specialized medical translator for English.  Translate
  the following clinical case report from Brazilian Portuguese to
  English.

  \medskip\noindent\textbf{Rules:}
  \begin{itemize}
    \item Preserve all medical terminology accurately.
    \item Maintain the original structure (section headings,
          paragraph breaks, markdown formatting).
    \item Keep numerical values, measurements, drug names, and
          abbreviations intact.
    \item Do not add or remove information.
    \item Return only the translated text, without explanations.
  \end{itemize}

  \medskip\noindent\textbf{Input format:}
\begin{verbatim}
## Title
[article title]

## Full Text
[full article text]
\end{verbatim}

  \noindent\textbf{Output format (follow exactly):}
\begin{verbatim}
## Title
[translated title]

## Full Text
[translated full text]
\end{verbatim}
\end{promptbox}

\begin{promptbox}
  \label{prompt18:en_pt_translation}
  \textbf{Prompt 18. Prompt for English-to-Portuguese translation of
  full-text case reports.}

  \medskip
  You are a medical translator specialized in Brazilian Portuguese
  (pt-BR).  Translate the following clinical case report from English
  to Brazilian Portuguese.

  \medskip\noindent\textbf{Rules:}
  \begin{itemize}
    \item Preserve all medical terminology with precision.
    \item Maintain the original structure (section headings,
          paragraph breaks, markdown formatting).
    \item Keep numerical values, measurements, drug names, and
          abbreviations intact.
    \item Do not add or remove information.
    \item Return only the translated text, without explanations.
  \end{itemize}

  \medskip\noindent\textbf{Input format:}
\begin{verbatim}
## Title
[article title]

## Full Text
[full article text]
\end{verbatim}

  \noindent\textbf{Output format (follow exactly):}
\begin{verbatim}
## Title
[translated title]

## Full Text
[translated full text]
\end{verbatim}
\end{promptbox}

\begin{promptbox}
  \label{prompt:case_classifier}
  \textbf{Prompt X. Prompt for classifying clinical case reports as
  single or multiple scenarios.}
  \medskip

  You are an experienced physician acting as a judge for a clinical
  reasoning benchmark. I will provide you with the complete text of a
  case report (it may be in Portuguese or English). Your task is to
  decide whether this report contains \textbf{a single clinical
  scenario} for evaluation or \textbf{multiple clinically distinct
  scenarios}.

  \medskip\noindent\textbf{Criterion:}
  \medskip

  \noindent Answer \textbf{``ONE''} if the report fits \textit{any}
  of the following situations:
  \begin{itemize}
    \item Describes a single patient.
    \item Describes multiple patients who essentially share the same
          diagnosis, the same investigation (tests, findings), and the
          same clinical reasoning. Since they are redundant for
          evaluating a model, they count as a single case.
    \item Example: ``two siblings with the same genetic mutation, same
          presentation, and same diagnosis'' $\rightarrow$ ONE.
    \item Example: ``series of 8 patients with diabetes treated with
          the same protocol, all with the same outcome''
          $\rightarrow$ ONE.
  \end{itemize}

  \noindent Answer \textbf{``MULTIPLE''} if the report describes two
  or more cases with clinically distinct presentations, diagnoses, or
  investigations, such that evaluating a model would require
  independent reasoning.
  \begin{itemize}
    \item Example: ``Case 1: uveal melanoma. Case 2: intraocular
          lymphoma'' $\rightarrow$ MULTIPLE.
    \item Example: ``Patient 1 presented with symptom A and received
          diagnosis X; Patient 2 presented with symptom B and received
          diagnosis Y'' $\rightarrow$ MULTIPLE.
    \item Example: ``series of 5 cases, each with a different
          diagnosis'' $\rightarrow$ MULTIPLE.
  \end{itemize}

  \medskip\noindent\textbf{Attention:}
  \begin{itemize}
    \item Do \textbf{not} use ``case series'' in the title as a
          definitive criterion. Evaluate the content.
    \item Mentions of other cohorts in the discussion or references
          (e.g., ``in previous studies with 30 patients\ldots'') do
          \textbf{not} make the report multiple.
  \end{itemize}

  \medskip\noindent\textbf{Response format:}

  \noindent Respond only in JSON, without additional text:
\begin{verbatim}
{
  "cases": "ONE" | "MULTIPLE",
  "justification": "<one sentence explaining your decision,
                    citing what determined your answer>"
}
\end{verbatim}

  \noindent\textbf{Input:}
\begin{verbatim}
# Complete text of the case report
{full_text}
\end{verbatim}
\end{promptbox}
\begin{promptbox}
  \label{prompt:differential_diagnosis}
  \textbf{Prompt. Prompt for differential diagnosis summarization of clinical case reports.}
  \medskip
  As an experienced physician, you will receive a clinical case report
  focused on diagnosis. Your task is to:
  \begin{itemize}
    \item Summarize the differential diagnosis process, including the
          rationale for each step and the reasons for considering or
          excluding specific diagnoses.
    \item Summarize the patient's final diagnosis.
  \end{itemize}
  Ensure that your summaries are concise and accurate, based exclusively
  on the information provided in the case report. Please avoid referencing
  any images, tables, or other visual data (if any). If the case report is
  incomplete or does not meet the requirements for producing a summary,
  simply respond: ``I cannot.''

  \medskip\noindent\textbf{Differential Diagnosis Process:}
  \medskip

  First, based on the patient's chief complaint and clinical information,
  an initial list of possible diseases should be generated. This list will
  then be systematically evaluated, comparing each disease against the
  patient's symptoms, signs, and test results.

  Next, the rationale for including or excluding each diagnosis will be
  explained based on clinical evidence, highlighting the selection of
  diagnostic tests that help confirm or rule out specific conditions. The
  process will involve the step-by-step elimination of less likely
  diagnoses, culminating in the most probable diagnosis, which will be
  justified by clinical reasoning and the supporting evidence.

  \medskip\noindent\textbf{Final Diagnosis (with explanatory details):}
  \medskip

  Integrate the patient's clinical presentation, test results, and
  differential diagnosis process to summarize the final diagnosis. Briefly
  explain the basis of the diagnosis and highlight the key factors
  supporting this conclusion.

  \medskip\noindent\textbf{Diagnosis Result:}
  \medskip

  Present only the diagnostic result, without any further explanation.

  \medskip\noindent\textbf{Output format (follow exactly):}
\begin{verbatim}
After the free-text reasoning, finalize your response with a JSON
block in exactly this format:

{
  "differential_diagnoses": [
    {"diagnosis": "...", "included": true,  "rationale": "..."},
    {"diagnosis": "...", "included": false, "rationale": "..."}
  ],
  "final_diagnosis_explanatory": "...",
  "diagnosis_result": "..."
}
\end{verbatim}
  \noindent\textbf{Input format:}
\begin{verbatim}
## Clinical Case
{case_text}
\end{verbatim}
\end{promptbox}

\end{document}